\documentclass[sigconf, nonacm]{acmart}

\newcommand\vldbdoi{XX.XX/XXX.XX}
\newcommand\vldbpages{XXX-XXX}
\newcommand\vldbvolume{18}
\newcommand\vldbissue{2}
\newcommand\vldbyear{2025}
\newcommand\vldbauthors{\authors}
\newcommand\vldbtitle{\shorttitle} 
\newcommand\vldbavailabilityurl{https://github.com/Yang-yuxin/BenchTGNN}
\newcommand\vldbpagestyle{plain} 

\usepackage{microtype}
\usepackage{graphicx}
\usepackage{subfigure}
\usepackage{booktabs} 
\usepackage{ragged2e}
\usepackage{blindtext}

\usepackage[ruled, lined, linesnumbered, commentsnumbered, longend]{algorithm2e}

\usepackage{cleveref}
\usepackage{amsmath,amsfonts}
\usepackage{enumerate}
\usepackage{enumitem}
\usepackage{hyperref}

\usepackage{array}
\usepackage{makecell}

\usepackage{multirow,float}
\usepackage{bm}   
\usepackage{xcolor}
\usepackage{pifont}

\usepackage{tikz,pgfplots}
\usepackage{pgfmath}
\usetikzlibrary{pgfplots.groupplots}
\usetikzlibrary{patterns}
\usetikzlibrary{shapes,snakes}
\usetikzlibrary{matrix}
\usetikzlibrary{arrows.meta}
\usepgfplotslibrary{fillbetween}
\newcommand{\tikzsetnextfilename}[1]{\ignorespaces}

\usepackage{comment}
\usepackage{xcolor}

\newcommand{\newcontent}[1]{\textcolor{black}{#1}}
\newcommand{\shepherding}[1]{\textcolor{black}{#1}}

\newcommand\blfootnote[1]{%
  \begingroup
  \renewcommand\thefootnote{}\footnote{#1}%
  \addtocounter{footnote}{-1}%
  \endgroup
}

\begin{document}
\title{Towards Ideal Temporal Graph Neural Networks: \\ Evaluations and Conclusions after 10,000 GPU Hours
}

\author{Yuxin Yang}
\affiliation{%
  \institution{University of Southern California}
  \city{Los Angeles}
  \state{California}
  \postcode{90089}
}
\email{yyang393@usc.edu}

\author{Hongkuan Zhou}
\affiliation{%
  \institution{University of Southern California}
  \city{Los Angeles}
  \state{California}
  \postcode{90089}
}
\email{hongkuaz@usc.edu}

\author{Rajgopal Kannan}
\affiliation{%
  \institution{DEVCOM Army Research Office}
  \city{Los Angeles}
  \state{California}
  \postcode{90089}
}
\email{rajgopal.kannan.civ@army.mil}

\author{Viktor Prasanna}
\affiliation{%
  \institution{University of Southern California}
  \city{Los Angeles}
  \state{California}
  \postcode{90089}
}
\email{prasanna@usc.edu}

\begin{abstract}
Temporal Graph Neural Networks (TGNNs) have emerged as powerful tools for modeling dynamic interactions across various domains.
The design space of TGNNs is notably complex, given the unique challenges in runtime efficiency and scalability raised by the evolving nature of temporal graphs. 
We contend that many of the existing works on TGNN modeling inadequately explore the design space, leading to suboptimal designs.
Viewing TGNN models through a performance-focused lens often obstructs a deeper understanding of the advantages and disadvantages of each technique. Specifically, benchmarking efforts inherently evaluate models in their original designs and implementations, resulting in unclear accuracy comparisons and misleading runtime.
To address these shortcomings, we propose a practical comparative evaluation framework that performs a design space search across well-known TGNN modules based on a unified, optimized code implementation. 
Using our framework, we make the first efforts towards addressing three critical questions in TGNN design, spending over 10,000 GPU hours: (1) investigating the efficiency of TGNN module designs, (2) analyzing how the effectiveness of these modules correlates with dataset patterns, and (3) exploring the interplay between multiple modules. 
Key outcomes of this directed investigative approach include demonstrating that the most recent neighbor sampling and attention aggregator outperform uniform neighbor sampling and MLP-Mixer aggregator; Assessing static node memory as an effective node memory alternative, and showing that the choice between static or dynamic node memory should be based on the repetition patterns in the dataset. Our in-depth analysis of the interplay between TGNN modules and dataset patterns should provide a deeper insight into TGNN performance along with potential research directions for designing more general and effective TGNNs.

\end{abstract}

\maketitle

\pagestyle{\vldbpagestyle}
\begingroup\small\noindent\raggedright\textbf{PVLDB Reference Format:}\\
\vldbauthors. \vldbtitle. PVLDB, \vldbvolume(\vldbissue): \vldbpages, \vldbyear.\\
\href{https://doi.org/\vldbdoi}{doi:\vldbdoi}
\endgroup
\begingroup
\renewcommand\thefootnote{}\footnote{\noindent
This work is licensed under the Creative Commons BY-NC-ND 4.0 International License. Visit \url{https://creativecommons.org/licenses/by-nc-nd/4.0/} to view a copy of this license. For any use beyond those covered by this license, obtain permission by emailing \href{mailto:info@vldb.org}{info@vldb.org}. Copyright is held by the owner/author(s). Publication rights licensed to the VLDB Endowment. \\
\raggedright Proceedings of the VLDB Endowment, Vol. \vldbvolume, No. \vldbissue\ %
ISSN 2150-8097. \\
\href{https://doi.org/\vldbdoi}{doi:\vldbdoi} \\
}\addtocounter{footnote}{-1}\endgroup

\ifdefempty{\vldbavailabilityurl}{}{
\vspace{.3cm}
\begingroup\small\noindent\raggedright\textbf{PVLDB Artifact Availability:}\\
The source code, dat
a, and/or other artifacts have been made available at \url{\vldbavailabilityurl}.
\endgroup
}
\section{Introduction} \label{sec:intro}

\blfootnote{\textbf{Distribution Statement A:} Approved for public release. Distribution is unlimited.}

Temporal graphs are a prevalent data type modeling numerous real-world systems of interactions and relations spanning from social networks~\cite{min2021stgsn, bai2020temporal}, economics~\cite{wang2021temporal, li2022corporate} to traffic networks~\cite{wang2020traffic, li2021spatial, bui2022spatial}. The complexity of temporal graphs stems from \shepherding{the evolving contextual and structural information}, as nodes and edges change over time. 
Dynamic Graph Representation Learning (DGRL) aims at mapping temporal graphs into low-dimensional spaces, that facilitates many downstream applications in forecasting~\cite{cao2020spectral, li2021spatial, 9636143}, recommendations~\cite{bai2020temporal, zhou2021temporal, chang2021sequential}, and anomaly detection~\cite{cai2021structural, zeng2021hierarchical, chen2022graphad}.

\begin{table*}[t]
    \setlength{\tabcolsep}{0.9mm}
    \centering
    \caption{Summary of module performance according to our experimental results. We denote the modules with consistently better performance regardless of varying combinations of other modules as \textit{more effective}, and their opposite as \textit{less effective}. \textit{Dataset dependent} represents the modules whose performance depends on datasets.}
    \begin{tabular}{r|ccc}
        \toprule
        Performance & Modules \\
        \midrule
        More effective & Attention aggregator, most recent neighbor sampling, non-learnable time encoding \\
        Dataset dependent & RNN-based node memory, embedding table-based node memory \\
        Less effective & MLP-Mixer aggregator, uniform neighbor sampling, learnable time encoding \\
        \bottomrule
    \end{tabular}
    \label{tab:summ}
\end{table*}

Among DGRL methods, Temporal Graph Neural Networks (TGNNs) stand out as an effective solution. 
\newcontent{Temporal graphs are divided by the continuity of the timestamps into two categories: Discrete Time Dynamic Graphs (DTDGs) and Continuous Time Dynmic Graphs (CTDGs)~\cite{kazemi2020representation}. DTDG methods integrate information from a set of static graph snapshots ~\cite{roland,yang2021discrete,evolvegcn,gao2022equivalence}, while CTDG methods aggregate information from temporal neighborhoods often obtained by sampling ~\cite{tgn,tgat,jodie,dyrep,dysat,apan,graphmixer,cawn,c-saw,neurtw}. We focus our discussion on CTDGs since DTDGs are a specific case of the more general CTDGs~\cite{tgl}.}
\newcontent{On CTDGs,} TGNNs generate dynamic node embeddings by performing time-aware message passing, arising from the well-established message-passing paradigms of static GNNs. 
\newcontent{The interleaved structural and temporal information poses challenges. At the same time,} runtime problems arise from the temporal nature of the graphs, such as the proportionally growing size of the temporal neighborhood with time and the time-dependent sequentially aggregated node features. 

Initial research efforts in TGNNs centered around the development of effective and efficient models~\cite{tgn, tgat, jodie, neurtw, taser, astgn, zebra, tgl, disttgl, gnnflow}.
Recently, a new stream of research has emerged, focusing on the critical evaluation and fair comparison of TGNN models~\cite{edgebank, tgb, benchtemp}. 
We identify two critical issues in this emerging research. \textbf{First}, the evaluation of TGNNs often lacks a comprehensive exploration of the design space. Newer TGNN models typically share similar sets of design choices with previous works, encompassing several {\it common modules}, with multiple options for each. However, existing works on model advancement pay little attention to searching the design space and are satisfied with incomplete success in limited spaces. Consequently, module advancements may not bring out the full extent of model performance when other parts of the architecture are carelessly designed. 
\textbf{Second}, model comparisons are not performed in a unified framework with reasonable optimization. 
\newcontent{For example, some works~\cite{chen2022bottleneck, tgb, tgn} carry out sequential neighbor sampling, while other works use parallel sampling ~\cite{taser}; TGN ~\cite{tgn} uses 
linear probing to identify the last message in a batch, while TGL~\cite{tgl} leverages 
GPU for parallel processing.}
The unoptimized code frameworks obscure accurate performance evaluation and create a distorted view of the comparative efficiency of different modules. As a result, there is a limited understanding of TGNN design space, hindering the development of truly optimized models.

In addressing these gaps, our paper proposes a practical approach to TGNN model comparison that compares models at the modular level with a standardized and optimized implementation framework. \newcontent{Unlike existing TGNN frameworks that highlight a complete pipeline ~\cite{dyglib, tgb} or distributed acceleration ~\cite{tgl}, our methodology is developed and optimized for evaluating individual modules to discern their contribution to the overall model efficacy.} Our comparison work focuses on achieving optimality where high predictive accuracy and resource efficiency are balanced. Spending over 10,000 GPU hours (NVIDIA RTX 6000 Ada) on seven datasets, we explore how module choices, sampling strategies, and dataset patterns influence optimality.
Specifically, we make the first efforts to address three key {\it interrelated} research questions in this work:


\noindent\textbf{RQ1. Efficiency and Cost-effectiveness of Module Designs.} 
What module designs, such as the choice between lightweight and more complex neighbor aggregators, strike the optimal balance between effectiveness and resource efficiency? 

\noindent\textbf{RQ2. Universality of Module Effectiveness across Datasets.}
Do the best-performing modules on some datasets, like specific types of node memory, maintain their effectiveness universally across all datasets, or is module performance dependent on dataset characteristics?

\noindent\textbf{RQ3. Interplay between Different Modules in the Model.} 
Does the integration of various modules enhance or undermine performance? For example, does the combination of node memory and a deeper neighbor sampling strategy integrate the improvements offered by each module?

Through extensive experiments based on a modular comparison framework, we obtain some initial answers to the \textbf{RQs}. We highlight some of the more important findings below with more detailed findings discussed subsequently. 
1. From the cost-effectiveness perspective, {\it most recent} neighbor samplers are better than {\it uniform} neighbor samplers and {\it attention-based} neighbor aggregators are better than the recently proposed {\it MLP-Mixer} neighbor aggregators. 2. We propose static node memory as a powerful node memory. Further comparison with RNN-based node memory shows that the effectiveness of node memory is largely influenced by dataset repetition patterns, which we also verify using synthetic datasets. 3. Combining node memory with a deeper neighbor sampling strategy proves ineffective due to their overlapping effects. We summarize results about TGNN module effectiveness in Table~\ref{tab:summ}. Our experimental results align with previous theoretical works, while also providing an enlightening perspective on datasets. Meaningful conclusions about TGNN modules are drawn from the results, leading to numerous research directions.

The paper is structured as follows. In Section~\ref{sec:bg}, we introduce several widely-used TGNN models, examine their common architectures, 
and develop a unified modular framework for evaluating such TGNN models.
In Section~\ref{sec:methods}, we detail the implementation of our modular comparison framework and experimental settings. In Section~\ref{sec:res}, we analyze how various modules and dataset characteristics impact model performance and optimality. In Section~\ref{sec:discuss}, we answer the research questions above based on experimental results. Finally, in Section~\ref{sec:future}, we summarize our findings and reveal their implications for future research.

\section{Background} \label{sec:bg}

Before delving into our experimental design and findings, we review popular TGNN models and existing comparative analyses. 
\newcontent{We reveal the extensive shared components across different models and identify two crucial issues arising from the lack of modular comparison frameworks:}
(1) TGNN models are making modular advancements on suboptimal model frameworks, leading to possible flaws in the advancements claimed. (2) Many current models use unoptimized code, \shepherding{which heavily hinders a fair comparison.} 

\subsection{Temporal Graph Neural Networks}

Temporal graphs can be interpreted as a set of timestamped interactions among nodes. Continuous time temporal graphs are usually represented as $\mathcal{G}(\mathcal{V},\mathcal{E})$ with events $\{(u, v, \bm {e}_{uvt}, t)\}$, where each quadruplet represents an interaction at time $t$ from node $u$ to node $v$ with feature $\bm {e}_{uvt}$. TGNNs leverage structural and temporal information in a temporal graph to generate informative low-dimensional node embeddings for prediction tasks. Like static GNNs, TGNNs utilize a message-passing framework to iteratively aggregate information from every node's temporal neighborhood. 

In this work, we mainly compare TGNN models that leverage an update-sampling-aggregation scheme as they form the major stream of TGNN models. 
These models share a general pipeline that aggregates information with an evolving timestamp: Given the timestamp for prediction, the model first updates the temporal neighborhoods and memories of the nodes. Next, past node-node interactions are sampled as neighbors within a node's temporal neighborhood. Finally, an aggregator combines the neighbors, node features, and node memories into an embedding \shepherding{ used for prediction.} 

This entire process can be encapsulated using three key modules: 
neighbor sampling, neighbor aggregation, and node memory. 
As illustrated in Table ~\ref{tab:models}, many widely-used TGNN models can be described using this {\it unified} and {\it modular} framework.

\begin{table}[h]
    \setlength{\tabcolsep}{0.1mm} 
    \renewcommand{\arraystretch}{0.9} 
    \small 
    \centering
    \caption{Design choices of different TGNN models. \textit{Neighbor Sampling}, \textit{Node Memory}, \textit{Neighbor Aggregator} columns contain the choices of model components. \textit{Add-on} denotes the add-on designs of corresponding models. MR: Most Recent neighbor sampling. RNN: Recurrent Neural Networks (including GRU, LSTM).}
    \begin{tabular}{c|cccc}
        \toprule
         Model & \begin{tabular}{c}\textit{Neighbor}\\\textit{Sampling}\end{tabular} 
         & \begin{tabular}{c}\textit{Node}\\ \textit{Memory}\end{tabular} 
         & \begin{tabular}{c}\textit{Neighbor}\\ \textit{Aggregator}\end{tabular}
         & \textit{Add-on}\\
        \midrule
        TGN~\cite{tgn} & MR & RNN & Attention & Learnable time encoding~\cite{time2vec} \\
        TGAT~\cite{tgat} & Uniform & - & Attention & Learnable time encoding\\
        APAN~\cite{apan} & MR & RNN & Attention & Learnable time encoding \\
        GraphMixer~\cite{graphmixer} & MR & - & MLP-Mixer & Fixed time encoding \\
        JODIE~\cite{jodie} & MR & RNN & Attention & Embedding projection \\
         & & & & one-hot identifier \\
         & & & & Learnable time encoding \\
        \bottomrule
    \end{tabular}
    \label{tab:models}
\end{table}

Note that random walk-based models ~\cite{cawn, neurtw, catwalk} 
utilize traditional graph analytic algorithms to boost the performance of TGNNs, thus we do not discuss them in this work.


\subsection{A modular perspective of TGNN models}\label{subsec:modular}


We elaborate on our framework by 
\shepherding{formalizing every module.}

\noindent \textbf{Neighbor sampling} \label{sec:sampler}
TGNNs use samplers to sample past interactions of a node for aggregation. For node $v$ at time $t$, a neighbor sampler uses a strategy to sample within its temporal neighborhood 
$\mathcal{N}(v, t_0) = \{(v, u, \bm e_{uvt}, t) \in \mathcal{E}, t < t_0\}$. 
Existing works often employed the most recent sampling ~\cite{tgn} or uniform sampling ~\cite{tgat}. In most recent sampling, a given number of most recent neighbors are sampled prior to the current timestamp. In uniform sampling, neighbors are uniformly sampled within all past interactions. 
\newcontent{Previous works also tested inverse time-span sampling, which performs worse than uniform sampling~\cite{tgat}. We do not include the adaptive sampling methods~\cite{taser, astgn} due to the tremendous computation and communication overheads introduced. }

\noindent \textbf{Node memory} \label{sec:mem}
Node memory is a module that stores information from a node's past interactions. They were usually implemented with a fixed-length embedding updated with interactions by sequential models ~\cite{tgn, jodie}. For an event $(u, v, \boldsymbol{e}_{uvt}, t)$, two messages are generated and sent to the mailboxes of node $u$ and node $v$:
\begin{align}
    \mathbf m_u&=\left\{\mathbf s_u||\mathbf s_v||\Phi(t-t^-_u)||\mathbf e_{uvt}\right\}\\
    \mathbf m_v&=\left\{\mathbf s_v||\mathbf s_u||\Phi(t-t^-_v)||\mathbf e_{uvt}\right\},
\end{align}
Here $\Phi(\cdot)$ is the time encoding function, $t^-_u$ is the timestamp of the last update, $s_u$ and $s_v$ are the latest node memory for nodes $u$ and $v$, and $\mathbf e_{uvt}$ is the edge feature representing the interaction. Then, we use an update function $\textup{UPDT}$, usually a Recurrent Neural Network (RNN) or its variant, to update the node memory of \shepherding{$u$ and $v$,}
\begin{align}
    \mathbf s_u=\textup{UPDT}\left(\mathbf s_u,\mathbf m_e^{uv}\right)\qquad \mathbf s_v=\textup{UPDT}\left(\mathbf s_v,\mathbf m_e^{vu}\right)\label{eq:memupdt}
\end{align}

In batch training, interactions with node $v$ in the same batch are usually reduced to the mean or the last interaction of the whole batch for efficiency. 

To reduce the costs of memory updates, some work~\cite{EDGE} leveraged a learnable static node embedding to capture the information of active nodes, bringing more flexibility to training and expediting inference. 
Some works also used a mixture of static and dynamic node embeddings ~\cite{disttgl, EDGE}. 

\noindent \textbf{Neighbor aggregator} \label{sec:aggr}
Neighbor aggregators are stacked to build a multi-layer TGNN model. 
\newcontent{It produces the embedding of any node $v$ at time $t$ by aggregating the embedding vectors of the sampled neighbors.}
Attention aggregator has been a popular choice among neighbor aggregators ~\cite{tgn, tgat, apan, jodie}.
Let $\boldsymbol{h}_v^{(l-1)}$ be the $(l-1)$-layer embedding of node $v$ (generated by previous layers), attention aggregator performs attention-based aggregation within its $(l-1)$-layer neighborhood:
\begin{align}       
    \boldsymbol{h}_v^{(l)}&= \operatorname{Atten} (\boldsymbol{h}_v^{(l-1)}, \boldsymbol{M}_v^{(l)}) \label{eq:tgat_fw}
\end{align}
where $\boldsymbol{M}_v^{(l)}$ is the $l$-th layer message matrix of temporal neighborhood $\mathcal{N}(v,t)$. 
In addition to the commonly used attention aggregator, a recent work, GraphMixer~\cite{graphmixer}, claimed that simply stacking feed-forward and convolution layers yields good node embedding.  
\begin{align}
            \boldsymbol{h}_v^{(l)} = 
            \operatorname{Mean}\left(\operatorname{MLP-Mixer}\left(\boldsymbol{M}_u^{(l)}
            \right)\right).
\end{align}

\newcontent{Theoretically, the complexity of MLP-Mixer~\cite{mlpmixer} is linear in the number of inputs, while the attention aggregator has a quadratic complexity. As TGNNs typically do not encode temporal neighborhoods with excessively long sequences, the complexity of using these two modules is comparable in practical experiments.}

\noindent \textbf{Other modules} \label{sec:add-on}
There are several powerful add-on modules in TGNN models, including time encoding ~\cite{tgn, tgat, graphmixer}, neighborhood representations ~\cite{nat}, time-lapse noise ~\cite{jodie}, etc. These modules usually produce embeddings as augmentations \shepherding{to existing features} at preprocessing, postprocessing, or aggregation stages. We briefly discuss the widely employed time encoding module and leave out the other model-specific modules.

\subsection{A unified model framework for comparison and benchmarking}

\begin{figure*}[t]
  \vspace{-30pt}
  \centering
  \setlength{\abovecaptionskip}{5pt} 
  \setlength{\belowcaptionskip}{-3pt} 
  \input{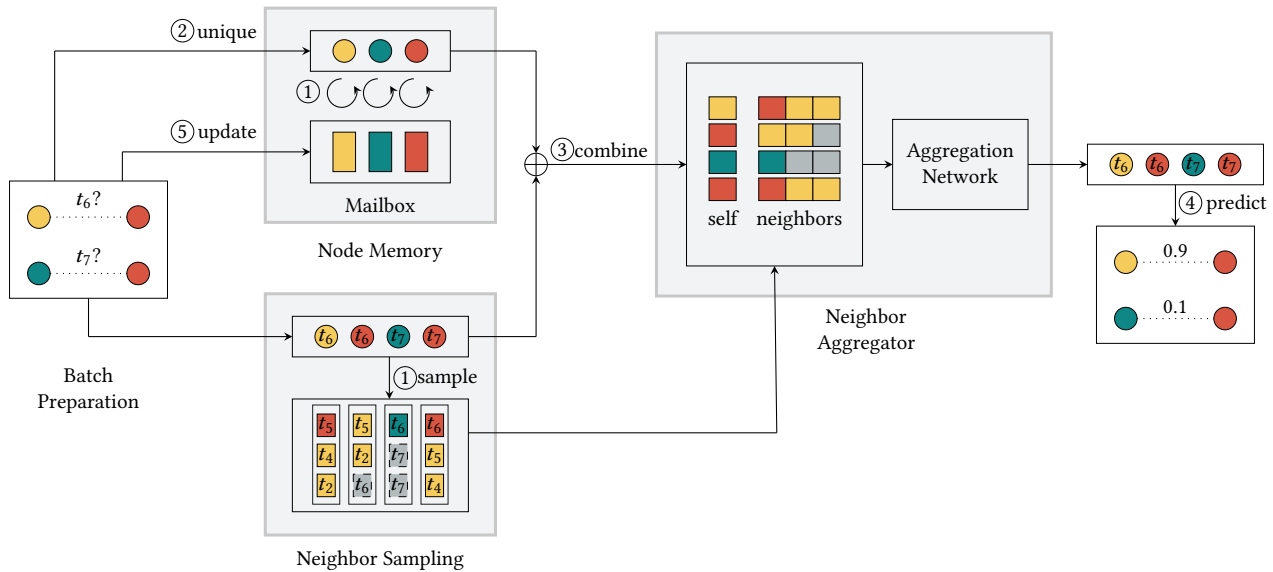}
  \caption{Overview of our generalized update-sampling-aggregation TGNN pipeline. 
  We focus our discussion on the three modules in grey rectangles: node memory, neighbor sampling, and neighbor aggregator. We use small circles to represent nodes and rectangles to represent edge embeddings. Arrows indicate the flow of data, encompassing transformations and rearrangements. In the neighbor sampling module, dotted dark grey rectangles represent dummy nodes where adequate valid neighbors are unavailable. The circled numbers illustrate the sequence of task execution, determined by task dependencies. The same numbers represent that the tasks can be executed parallelly without dependency issues.}
  \label{fig:framework}
\end{figure*}

Despite the numerous shared design choices, existing works often use different implementations for the same module, some of which involve unoptimized code. For example, TGN implementation involves a time-consuming memory updating stage with slow linear probing to find the last message in a batch. This occupies almost 50\% of total execution time where batch size $bs = 600$. However, this time may be reduced by a factor of ten in our code.
These unoptimized implementations hinder a fair comparison of the runtime and efficiency of those modules, consequently obstructing insights into the cost-effectiveness of TGNN modules. We aim to bridge this gap by evaluating the effectiveness of these modules under a unified and optimized framework (See Figure ~\ref{fig:framework}). To the best of our knowledge, no works provided a fair comparison of these models at such granularity. We also use an improved metric (Mean Reciprocal Rank, MRR) and include a more challenging inductive setting, following some recent benchmarking works ~\cite{benchtemp, tgb}.

\section{Methodology} \label{sec:methods}

\shepherding {In this section, we describe our experimental framework, including the tasks, metrics, and validation framework for comparison.}

\subsection{Tasks and Metrics} \label{subsec:task}

We evaluate the models with the link prediction tasks on continuous-time dynamic graph datasets~\cite{jodie, opsahl2009clustering, panzarasa2009patterns, edgebank}. \newcontent{The datasets for node classification tasks are not well-established, with poor and biased label quality in existing datasets like \textit{Wikipedia} and \textit{REDDIT}. Temporal graph classification has been introduced recently~\cite{tpgnn}. Additionally, the lack of supervision in node classification tasks has led to a challenging two-stage training pipeline that is difficult to optimize or evaluate. We focus our evaluation on the self-supervised link prediction task, which generates informative embeddings with good performance in downstream tasks~\cite{tgat, tgn, tgl}.} We use Mean Reciprocal Rank (MRR) as the performance metric ~\cite{tgb, 10.14778/3551793.3551831}, for its enhanced capability to capture the ranking of the positive edge among the mixed positive and negative edges~\cite{tgb}.

In the transductive setting, we directly train and test our model on these sets. In the inductive setting, we mask 10\% of the nodes from test sets as unseen nodes and remove all edges with these unseen nodes from the training set. The edges containing the unseen nodes are considered inductive samples in the test set.
We sample 1:1, 1:9, and 1:49 positive and negative edges on train, validation, and test sets. For negative edge sampling within bipartite graphs, we only select destination nodes from the class different from that of the source node.

\subsection{Datasets}\label{subsec:datasets}

\begin{table}[b]
    \small
    \setlength{\tabcolsep}{0.9mm}
    \centering
    \setlength{\abovecaptionskip}{5pt} 
  \setlength{\belowcaptionskip}{-5pt} 
    \caption{Dataset Statistics. $|d_v|$ and $|d_e|$ denote the dimensions of node features and edge features, respectively. $\overline{k}$ denotes average node degree. Average degree is computed by $\frac{\# Used \ Edges}{\# Nodes}$. } \label{tab:ds}
    \begin{tabular}{r|cccccccc}
        \toprule
        & $|\mathcal{V}|$ & $|\mathcal{E}|$ & $|d_v|$ & $|d_e|$ & $\overline{k}$ & $\max(t)$ \\
        \midrule
        \textit{Wikipedia} & 9,227 & 157,474 & - & 172 & 17.07 & 2.7e6 \\
        \textit{Reddit} & 10,984 & 672,447 & - & 172 & 61.22 & 2.7e6 \\
        \textit{UCI} & 1,900 & 59,836 & - & - & 31.49 & 1.7e7\\
        \textit{CollegeMsg} & 1,900 & 59,835 & - & - & 31.49 & 1.1e9\\
        \textit{MOOC} & 7,144 & 411,749 & 172 & 4 & 57.64 & 2.6e6 \\
        \textit{LastFM} & 1,980 & 1,293,103 & 172 & 2 & 505.05 & 1.3e8 \\
        \textit{Flights} & 13,169 & 1,927,145 & 100 & - & 75.94 & 1.2e2\\
        \newcontent{\textit{GDELT}} & \newcontent{16,682} & \newcontent{191,290,882} & \newcontent{413} & \newcontent{130} & \newcontent{59.94} & \newcontent{1.6e8} \\
        \newcontent{\textit{SuperUser}} & \newcontent{194,086} & \newcontent{1,443,339} & \newcontent{128} & \newcontent{-} &  \newcontent{7.44} & \newcontent{2.4e8} \\
        \bottomrule
    \end{tabular}
\end{table}
We conduct our evaluation using a variety of real-world temporal graph datasets from diverse domains at all scales. 
\noindent The statistics of the datasets are shown in Table \ref{tab:ds}. 
The details of the temporal datasets are as follows:\\
\noindent\textit{Wikipedia}~\cite{jodie} is a bipartite graph of online edits. Editors and pages are nodes, and edges with timestamps represent the editing events. Edge features are extractions of the edited text content. \\
\noindent\textit{REDDIT}~\cite{jodie} is a bipartite graph of user posts. Users and subreddits are nodes, and edges are interactions as users post to the subreddits. Edge features are extracted from the texts of the posts. \\
\noindent\textit{MOOC}~\cite{jodie} is a bipartite online course content providing interaction network. Students and course units are nodes. Edges represent users interacting with a course unit. \\
\noindent\textit{LastFM}~\cite{jodie} is a bipartite user-song interaction network. Users and songs are nodes. Edges represent user listening events. \\
\noindent\textit{UCI}~\cite{opsahl2009clustering} is a user-forum social network. Nodes are students at the University of California, Irvine. Edges are interactions of online messages between users. 
\shepherding{Features are extracted from the messages. \\}
\noindent\textit{CollegeMsg}~\cite{panzarasa2009patterns} is a dataset derived from the social network introduced in dataset UCI. \\
\noindent\textit{Flights}~\cite{edgebank} is a \newcontent{transportation network of airline travels}. The nodes are airports, while the edges are flights. Edge features are the number of daily flights between two airports. \\
\noindent\newcontent{\textit{GDELT}~\cite{tgl}is a global event knowledge graph. Nodes represent actors, and edges represent events. The node and edge features are the CAMEO codes of actors and events.} \\
\noindent\newcontent{\textit{SuperUser}~\cite{orca} is a temporal network of interactions on the stack exchange website SuperUser. The nodes are users, and the edges represent different types of interactions.} 

\newcontent{The node features of \textit{LASTFM}, \textit{MOOC}, and \textit{SuperUser} are randomly generated following previous works~\cite{tgl, disttgl, orca}.}
\newcontent{For large datasets, we use the last 1 million edges following~\cite{taser} and split them chronologically into 70\%-15\%-15\% to build train, validation, and test sets. }

\subsection{Model implementation} \label{subsec:imple}

\newcontent{In this section, we present our experimental setting by describing the framework first and then the implementation of the neighbor sampler, memory module, and neighbor aggregator. 
As our framework is modular, it is relatively easy to incorporate and evaluate other methods.}

\noindent \newcontent{\textbf{TGNN framework.} Our TGNN framework (Figure~\ref{fig:framework}) consists of four stages: neighbor sampling, node memory update, neighbor aggregation, and prediction. After the neighbor sampler samples the edge batch node-wise from the temporal neighborhood, we update the node memory. Then, we use the neighbor aggregator to aggregate the sampled neighborhood for every node. Finally, a task-specific predictor is employed to generate predictions.}

\noindent\textbf{Neighbor sampler.} Neighbor sampling is a challenging task on dynamic graphs as node neighborhoods constantly evolve with timestamps. We employ the state-of-the-art GPU sampler presented in ~\cite{taser}. At the pre-processing stage, we sort all the edges and store the temporal neighbors of each node with a T-CSR data structure~\cite{tgl}, where neighbors of each node are arranged according to their timestamps. We perform a binary search to identify the timestamp-dependent neighborhood at a given timestamp. During the mini-batch fetching stage, newcontent{the edges are sampled in random or chronological order depending on the type of node memory.} We use a GPU sampler to parallelize the sampling process: We assign a block of threads to perform the sampling of the neighbors of a node at a given timestamp. By introducing a parallel neighbor sampler, we significantly reduce the sampling overhead in our model. Following~\cite{taser}, we set batch size $bs=600$ for the train set and $bs = 100$ for validation and test sets. \

\noindent\textbf{Memory module.} We implement sequentially updated memory modules and static node memory on GPU. For sequentially-updated memory modules, existing works~\cite{tgn, apan, jodie} usually maintained an up-to-date node memory on GPU. They use a Gated Recurrent Unit (GRU) as the node memory updater, with interactions encoded as fixed-length messages. It should be noted that RNN, GRU, or LSTM (Long short-term memory) produce comparable outcomes~\cite{tgn}, saving the effort of conducting repeated experiments across all node memory updaters. We collectively refer to this type of memory as RNN-based node memory. In RNN-based node memory, training and inference are accelerated using chronological batch training~\cite{tgn, apan, jodie, tgl}. Our batch processing is as follows: As in the state-of-the-art~\cite{tgl, apan}, we maintain a mailbox for every node. Messages sent to a node in the previous batches are stored in the mailbox, and node memory is updated before processing every batch. If a node receives multiple messages within a batch, the messages are usually reduced to the latest or average. We employ the former for simplification and efficiency, considering there is only a minor difference in performance~\cite{tgn}. For further acceleration, we retrieve the last message by parallelized batch operations on GPU, instead of linear probing as in some existing works~\cite{tgn, tgb, benchtemp}. As supervision comes from the current batch, we have to prevent information leakage from the current batch by updating the node memory with cached messages from the previous batch~\cite{tgn, apan, jodie}. At the end of each batch processing, we detach the updated node memory from the computation graph so that the depth of backward propagation on RNN parameters is controlled. 

\newcontent{For static learnable node memory, we use an embedding table with shape $[|\mathcal{V}|, D_{mem}]$ to store the node memory vectors, where $D_{mem}$ is the dimension of the static node memory.} The training set is shuffled in each epoch to generate mini-batches, in which the relevant node memory vectors are directly updated by the gradients. Note that this shuffled mini-batching does not apply to RNN-based memory, as it needs to be updated in chronological order. We follow ~\cite{tgl, disttgl, taser} and set the dimensions of all memory modules to 100, a reasonable length to achieve high performance without introducing too much computation.

\noindent\textbf{Neighbor aggregator.} We implement attention block~\cite{vaswani2017attention} and MLP-mixer~\cite{mlpmixer} as neighbor aggregators. \newcontent{Following existing works~\cite{tgn, tgl}, we combine raw node features, node memories, and time encodings into node embeddings by concatenation, linear transformations, and additions before neighbor aggregation.} The aggregators then perform feature aggregation and produce informative low-dimensional node embedding. For the attention aggregator, we set attention heads $h=2$ as in ~\cite{tgn, tgl, disttgl, taser}.

\noindent\textbf{Abbreviations.} We summarize all the modules and their abbreviations tested in our work.
\begin{itemize}
    \item \textit{Neighbor Sampler}: \textbf{MR} (most-recent neighbor sampler), \textbf{uni} (uniform neighbor sampler)
    \item \textit{Node Memory}: \textbf{RNN} (RNN-based node memory), \textbf{emb} (embedding table-based node memory)
    \item \textit{Neighbor Aggregator}: \textbf{atten} (attention neighbor aggregator), \textbf{mixer} (MLP-Mixer neighbor aggregator)
\end{itemize}

\subsection{Hardware and Software}\label{subsec:hs}

We implement our code on a machine with dual 96-core AMD EPYC 9654 CPUs with 1.5TB ECC-DDR5 RAM and NVIDIA RTX 6000 Ada GPU with 48GB ECC-GDDR6 VRAM.
Our experiments are performed in an environment with Python 3.9, PyTorch 2.1.0, and CUDA 11.8 on Ubuntu 22.04 LTS with Linux kernel version 5.15.0-100-generic.

\newcontent{To use our framework on multiple GPUs, parallel samplers can be launched on every GPU to sample in the temporal neighborhoods of nodes. The edge features, node features (if any), and the global copy of the node memory is stored in the shared memory. The gradients in each iteration are synchronized among the trainer processes through the NCCL backend. }

\section{Results} \label{sec:res}

After constructing the benchmarking framework, we run extensive experiments using over 10,000 GPU hours
and thoroughly analyze the results. We first separately examine the cost-effectiveness and universality of modules, including neighbor sampling (subsection ~\ref{subsec:ns}), memory module (subsection ~\ref{subsec:mem}), neighbor aggregator, and others (subsection~\ref{subsec:others}). Following this, we explore the effectiveness of these components when combined to construct the model.  We conclude by addressing the research questions posed in the introduction (Section~\ref{sec:intro}) based on our experimental outcomes.

\subsection{Neighbor Sampling} \label{subsec:ns}

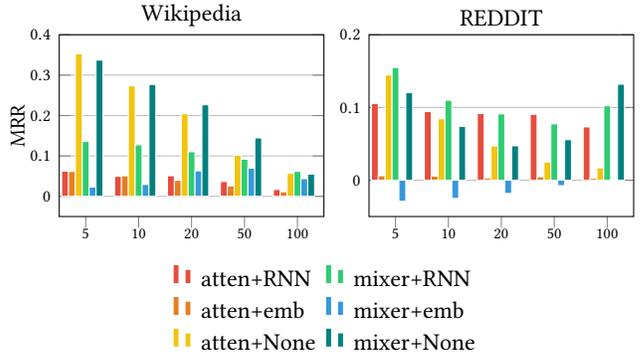
\begin{figure}[tb]
    \centering
    \setlength{\abovecaptionskip}{5pt} 
    \setlength{\belowcaptionskip}{-3pt} 
    \tikzsetnextfilename{sampling1}
\begin{tikzpicture}

    \definecolor{c1}{RGB}{231, 76, 60}
    \definecolor{c2}{RGB}{230, 126, 34}
    \definecolor{c3}{RGB}{241, 196, 15}
    \definecolor{c4}{RGB}{46, 204, 113}
    \definecolor{c5}{RGB}{52, 152, 219}
    \definecolor{c6}{RGB}{0, 128, 128}
        
    \begin{groupplot}[
        group style={
            group size=2 by 1,
            horizontal sep=0.6cm,
        },
        tick label style={font=\scriptsize},
        label style={font=\small},
        width=5.1cm,
        height=4cm,
    ]
    \nextgroupplot[ybar=0, bar width=2.6pt,title={Wikipedia},title style={yshift=-1.5ex},ylabel={MRR},ylabel style={yshift=-5ex},ymajorgrids,
        ymin=-0.05,ymax=0.4,xmin=0.5,xmax=5.5,xticklabel style={inner sep=0pt}, xtick pos=lower,
        scaled x ticks=false, xtick={1,2,3,4,5}, xticklabels={5,10,20,50,100}, ytick={0.0,0.1,0.2,0.3,0.4},
        legend pos=south east,legend style={legend columns=1,font=\fontsize{7}{8}\selectfont},legend image post style={xscale=0.75}]
        \addplot[style={white,fill=c1,mark=none}] table[col sep=comma, x expr=\coordindex+1, y=atten+RNN_WIKI]{tikz_data/two_ns.csv}; \label{l1}
        \addplot[style={white,fill=c2,mark=none}] table[col sep=comma, x expr=\coordindex+1, y=atten+emb_WIKI]{tikz_data/two_ns.csv}; \label{l2}
        \addplot[style={white,fill=c3,mark=none}] table[col sep=comma, x expr=\coordindex+1, y=atten+None_WIKI]{tikz_data/two_ns.csv}; \label{l3}
        \addplot[style={white,fill=c4,mark=none}] table[col sep=comma, x expr=\coordindex+1, y=mixer+RNN_WIKI]{tikz_data/two_ns.csv}; \label{l4}
        \addplot[style={white,fill=c5,mark=none}] table[col sep=comma, x expr=\coordindex+1, y=mixer+emb_WIKI]{tikz_data/two_ns.csv}; \label{l5}
        \addplot[style={white,fill=c6,mark=none}] table[col sep=comma, x expr=\coordindex+1, y=mixer+None_WIKI]{tikz_data/two_ns.csv}; \label{l6}
        \addplot[black, dashed] coordinates {(0,0) (5.5,0)};
        
    \nextgroupplot[ybar=0, bar width=2.6pt,title={REDDIT},title style={yshift=-1.5ex},ylabel style={yshift=-5ex},ymajorgrids,
        ymin=-0.05,ymax=0.2,xmin=0.5,xmax=5.5,xticklabel style={inner sep=0pt}, xtick pos=lower,
        scaled x ticks=false, xtick={1,2,3,4,5}, xticklabels={5,10,20,50,100}, ytick={-0.2,-0.1,0.0,0.1,0.2,0.3,0.4},
        legend pos=south east,legend style={legend columns=1,font=\fontsize{7}{8}\selectfont},legend image post style={xscale=0.75}]
        \addplot[style={white,fill=c1,mark=none}] table[col sep=comma, x expr=\coordindex+1, y=atten+RNN_REDDIT]{tikz_data/two_ns.csv};
        \addplot[style={white,fill=c2,mark=none}] table[col sep=comma, x expr=\coordindex+1, y=atten+emb_REDDIT]{tikz_data/two_ns.csv};
        \addplot[style={white,fill=c3,mark=none}] table[col sep=comma, x expr=\coordindex+1, y=atten+None_REDDIT]{tikz_data/two_ns.csv};
        \addplot[style={white,fill=c4,mark=none}] table[col sep=comma, x expr=\coordindex+1, y=mixer+RNN_REDDIT]{tikz_data/two_ns.csv};
        \addplot[style={white,fill=c5,mark=none}] table[col sep=comma, x expr=\coordindex+1, y=mixer+emb_REDDIT]{tikz_data/two_ns.csv};
        \addplot[style={white,fill=c6,mark=none}] table[col sep=comma, x expr=\coordindex+1, y=mixer+None_REDDIT]{tikz_data/two_ns.csv};
        \addplot [black, dashed] coordinates {(0,0) (5.5,0)};

    \end{groupplot}
    
    \matrix[matrix of nodes, anchor=north, inner sep=0.2em, draw=none, column 2/.style={anchor=base west},column 4/.style={anchor=base west}] at ([yshift=-0.5cm] group c1r1.south east)
    {
        \ref{l1} & atten+RNN   &
        \ref{l4} & mixer+RNN   \\
        \ref{l2} & atten+emb   &
        \ref{l5} & mixer+emb   \\
        \ref{l3} & atten+None   &
        \ref{l6} & mixer+None   \\
    };
\end{tikzpicture}%
    \caption{Performance gap between two neighbor samplers across various module combinations on datasets WIKI and \textit{REDDIT}. The values are calculated by subtracting the prediction accuracy of models using uniform sampling from the prediction accuracy of models using the most recent sampling. Results are plotted with varying numbers of sampled neighbors for 1-layer models.}
    \label{fig:sampling1}
\end{figure}

We first concentrate on neighbor sampling as sampling directly affects the information aggregated. Our analysis involves comparing two neighbor samplers, deciding the appropriate sampling budget, and designing a distribution strategy among layers given the total budget.
Specifically, we compare two popular samplers: \textbf{MR} and \textbf{uni} (See Figure~\ref{fig:sampling1}), \newcontent{both with little overhead ($1\%-2\%$ of training time) when carried out by our GPU sampler}. We observe that \textbf{MR} samplers often exceed \textbf{uni} by a large margin. \newcontent{When using \textbf{emb}, the performance gap is smaller, and sometimes \textbf{uni} sampler is better than \textbf{MR} sampler. This could be caused by the shared nature of static node embedding and uniform sampling of capturing long-term temporal neighborhoods, resulting in good training dynamics.} Due to space limits, we only display the results of 1-layer models on two datasets here. We observe similar phenomena in most experimental settings, indicating consistent superiority of \textbf{MR} neighbor sampler over \textbf{uni} neighbor sampler across various model settings and datasets. This result aligns with the preliminary results in ~\cite{zebra}. \newcontent{We observe a larger MRR improvement margin of \textbf{MR} over \textbf{uni} on \textit{Wikipedia} than \textit{REDDIT}, which may be explained by the recency of important events in the temporal neighborhood (further explained in Section ~\ref{subsec:mem}). As \textit{Wikipedia} is a dataset with short-term repetition patterns, \textbf{MR} may capture more relevant events on it, resulting in even better performance on some datasets.}

Next, we explore the effect of neighbor sampling budgets on TGNN model performance. We evaluate the performance of 1-layer and 2-layer models with different numbers of sampled neighbors in each layer. As most TGNNs incorporate layer sampling~\cite{tgn,tgat,jodie,dysat,dyrep,tgl,disttgl,taser}, the neighborhood size grows exponentially with layers (a problem known as neighbor explosion~\cite{sage}). Considering the high computational burden of layerwise sampling and the impact of a larger receptive field from node memory as discussed in~\cite{disttgl}, we only perform experiments on $\leq 2$ layers. 
\shepherding{For 1-layer models, we sampled 1 to 100 neighbors.}
For 2-layer models, we test combinations of 5 and 10 neighbors in each layer.

\begin{figure}[tb]
    \centering
    \setlength{\abovecaptionskip}{5pt} 
    \setlength{\belowcaptionskip}{-3pt} 
    \input{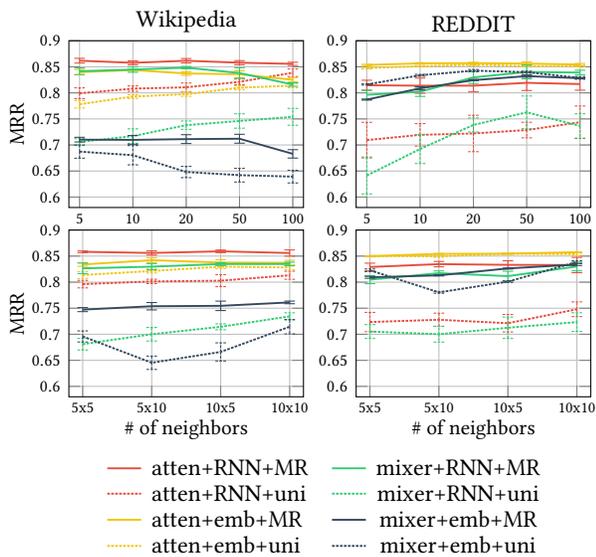}
    \caption{Performance of different combinations of TGNN modules on dataset \textit{Wikipedia} and \textit{REDDIT}. As more neighbors are sampled in models with one layer (first row) and two layers (second row), certain models display enhanced accuracy, while the top-performing model exhibits fast saturation.}
    \label{fig:sampling2}
\end{figure}

Our results indicate that TGNN models do not need many supporting neighbors or deep model architecture to perform well. Specifically, while some models improve as more neighbors are sampled, models with the highest-performing combinations reach their peaks quickly with less than 10 neighbors in 1-layer models(Figure ~\ref{fig:sampling2}). 
\shepherding{The result holds for all datasets.}
Combining past theoretical works~\cite{cong2024generalization} and our experimental results, we formulate a {\it three-fold hypothesis} concerning the impact of neighbor sampling on model performance: 
1. {\bf Sampling versus node memory}: The ability of node memory to capture the temporal neighborhood may be compromised due to discarded interactions in mini-batching or defective memory implementations. However, sampling can compensate for insufficient node memory and improve performance, i.e., sampling neighbors directly from a broader neighborhood may compensate for these deficiencies.
2. {\bf Deeper models and generalization gap}: Deeper models induce a generalization gap that contributes to performance saturation. We observe that on small-scale datasets like \textit{UCI} and \textit{CollegeMsg}, sampling more neighbors brings little to no performance improvement. Also, 2-layer models generally achieve better predictions at train time but similar predictions at test time compared to 1-layer models. These results align with the theoretical analyses in ~\cite{cong2024generalization} on how the generalization error of TGNNs decreases with data volumes and increases with depth (number of layers). 
3. {\bf Sampling overcomes suboptimality}: We observe that larger sampling sizes can improve the performance of suboptimal module combinations, alleviating module incompatibility. For instance, pairing \textbf{RNN} node memory with uniform sampling 
is suboptimal since it
fails to ensure that the node memory is updated with the latest interactions, potentially compromising the node memory's effectiveness; However increasing the sample size of uniform sampling imitates the reasonable \textbf{MR} sampling and memory update process, thus improving model performance, though the combination is suboptimal.


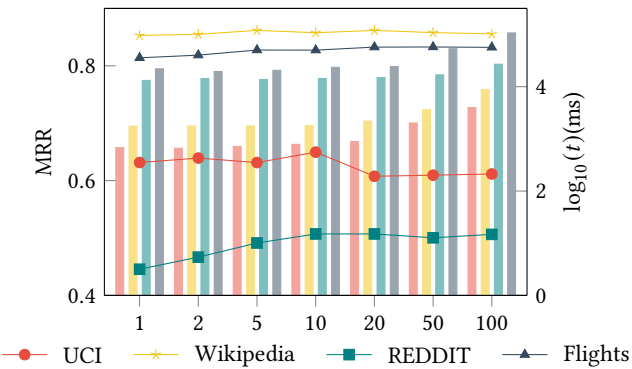
\begin{figure}[tb]
    \centering
    \setlength{\abovecaptionskip}{5pt} 
    \setlength{\belowcaptionskip}{-3pt} 
    \tikzsetnextfilename{mem_sat}
\begin{tikzpicture}

    \definecolor{c1}{RGB}{231, 76, 60}
    \definecolor{c2}{RGB}{241, 196, 15}
    \definecolor{c3}{RGB}{0, 128, 128}
    \definecolor{c4}{RGB}{52, 73, 94}

    \begin{axis}[
        ybar, 
        axis y line*=right, 
        xtick={1,2,3,4,5,6,7},
        xticklabels={1,2,5,10,20,50,100}, xtick pos=lower,
        ylabel={$\log_{10}(t) $(ms)}, ylabel near ticks, ylabel style={yshift=0ex},
        bar width=3pt,
        ymin=0, ymax=5.5, 
        width=7.2cm,
        height=5.4cm,
        name = ax1,
    ]
        \addplot[style={c1,fill=c1,mark=none, opacity=0.5}] table[col sep=comma, x expr=\coordindex+1, y=uci]{tikz_data/mem_sat_time.csv}; \label{bl1}
        \addplot[style={c2,fill=c2,mark=none, opacity=0.5}] table[col sep=comma, x expr=\coordindex+1, y=WIKI]{tikz_data/mem_sat_time.csv}; \label{bl2}
        \addplot[style={c3,fill=c3,mark=none, opacity=0.5}] table[col sep=comma, x expr=\coordindex+1, y=LASTFM]{tikz_data/mem_sat_time.csv}; \label{bl3}
        \addplot[style={c4,fill=c4,mark=none, opacity=0.5}] table[col sep=comma, x expr=\coordindex+1, y=Flights]{tikz_data/mem_sat_time.csv}; \label{bl4}
        
    \end{axis}

    \begin{axis}[
        axis y line*=left, 
        xtick={1,2,3,4,5,6,7},
        xticklabels={}, 
        xtick style={draw=none}, 
        ylabel={MRR}, ylabel near ticks, ylabel style={yshift=-0ex},
        ymin=0.4, ymax=0.9, 
        width=7.2cm,
        height=5.4cm,
    ]
        \addplot[sharp plot,style={c1, mark=otimes*, mark color=c1}] table[col sep=comma, x expr=\coordindex+1, y=uci_mean]{tikz_data/mem_sat_mrr.csv}; \label{ll1}
        \addplot[sharp plot,style={c2, mark=star, mark color=c2}] table[col sep=comma, x expr=\coordindex+1, y=WIKI_mean]{tikz_data/mem_sat_mrr.csv}; \label{ll2}
        \addplot[sharp plot,style={c3, mark=square*, mark color=c3}] table[col sep=comma, x expr=\coordindex+1, y=LASTFM_mean]{tikz_data/mem_sat_mrr.csv}; \label{ll3}
        \addplot[sharp plot,style={c4, mark=triangle*, mark color=c4}] table[col sep=comma, x expr=\coordindex+1, y=Flights_mean]{tikz_data/mem_sat_mrr.csv}; \label{ll4}
    \end{axis}
    \matrix (legend) at ([yshift=-0.8cm] ax1.south) [matrix of nodes, nodes={anchor=west}] {
        \ref{ll1} & UCI & [2mm] &
        \ref{ll2} & Wikipedia & [2mm] &
        \ref{ll3} & REDDIT & [2mm] &
        \ref{ll4} & Flights \\
    };

\end{tikzpicture}%
    \caption{\shepherding{MRRs and runtimes of 1-layer models sampling an increasing number of neighbors on datasets \textit{UCI}, \textit{Wikipedia}, \textit{REDDIT}, and \textit{Flights}. We use \textbf{atten} neighbor aggregator and \textbf{MR} sampling on all datasets. For node memory, we adopt the better-performing ones on each dataset: \textbf{emb} memory on \textit{REDDIT} and \textit{Flights}, and \textbf{RNN} memory on other datasets.} MRR is shown by line chart and runtime is shown by bar chart in logarithmic scale. Note that the x-axis uses a scale that has been slightly adjusted to ensure visual tidiness. }
    \label{fig:mem_sat}
    
\end{figure}

Next, we discuss the cost-effectiveness of neighbor sampling. As the size of the sampled neighborhood grows, model performance follows a convex curve and saturates quickly, whereas the runtime increases linearly. We plot the convex performance curve of MRRs (Figure~\ref{fig:mem_sat}) on all datasets of different scales.
\shepherding{While the saturation point differs, 1-layer model with 10 neighbors is close to top-performing on all datasets.}
Using the best-performing module combination, performance saturates with less than 10 supporting neighbors in 1-layer models. On the other hand, the runtime grows in proportion to the size of supporting neighbors, which is illustrated by a linear trend on the plot with both axes in logarithmic scale.

Note that increasing the depth and width of neighbor sampling still yields large improvements when node memory is absent, as shown in~\cite{tgat}. This corresponds to our first hypothesis. We skip relevant discussions for two reasons: (1). Models without node memory are not the focus of our design space search. (2). We demonstrate in Section~\ref{subsec:mem} that without node memory, models with a large sampling budget and longer runtime still produce worse results than very shallow models with node memory. 

In summary, the most recent neighbor sampling strategy {\it performs better} than uniform sampling. \shepherding{In models with node memory,} deep or wide neighbor sampling brings minor improvements to prediction accuracy but substantially increases the runtime. We claim that a small neighbor sampling budget (e.g., 1-layer models with 10 neighbors) suffices to aggregate neighborhood information.

\subsection{Memory modules}\label{subsec:mem}

Carefully designed node memory can alleviate computational burdens by capturing the informative temporal neighborhood, thereby reducing the need for extensive neighbor sampling and aggregation. We analyze the performance of \textbf{RNN} and \textbf{emb} node memory and demonstrate how to select the appropriate node memory.
First, we show by empirical results that the choice of node memory significantly affects the model performance. Then, we propose a hypothesis on the connection between node memory and dataset repetition patterns. After that, we introduce several measures to analyze the repetition patterns qualitatively and quantitatively. We then reveal a relationship between the repetition patterns of datasets and the performance of node memories. Finally, we construct synthetic datasets with different repetition patterns and validate our hypothesis with the results on the synthetic datasets.

\subsubsection{Empirical results and hypothesis.}
\begin{figure}[tb]
    \centering
    \setlength{\abovecaptionskip}{5pt} 
    \setlength{\belowcaptionskip}{-3pt} 
    \tikzsetnextfilename{mrr_2mem}
\begin{tikzpicture}

    \definecolor{cWIKI}{RGB}{241, 196, 15}
    \definecolor{cREDDIT}{RGB}{230, 126, 34}
    \definecolor{cFlights}{RGB}{52, 73, 94}
    \definecolor{cLASTFM}{RGB}{0, 128, 128}
    \definecolor{cMOOC}{RGB}{52, 152, 219}
    \definecolor{cUCI}{RGB}{231, 76, 60}
    \definecolor{cCollegeMsg}{RGB}{46, 204, 113}
    \definecolor{cGDELT}{RGB}{155, 89, 182}
    \definecolor{cSuperUser}{RGB}{128, 128, 128}

    \begin{axis}[
        xlabel={MRR (\textbf{RNN})}, xtick={0.0,0.2,0.4,0.6,0.8,1.0}, xtick pos=lower, xlabel style={yshift=1ex},
        ylabel={MRR (\textbf{emb})}, ytick={0.0,0.2,0.4,0.6,0.8,1.0}, ylabel near ticks, ylabel style={yshift=-1ex}, ytick pos=left,
        ymin=0.0, ymax=1.0, 
        xmin=0.0, xmax=1.0, 
        width=4.5cm,
        height=4.5cm,
        name = ax1,
        legend style={
        at={(1.15,0.5)}, 
        anchor=west,
        legend cell align=left,
        draw=none, 
        font=\small, 
        column sep=7pt
        }
    ]

    \addplot [
            scatter,
            only marks,
            point meta=explicit symbolic,
            legend pos=outer north east,
            scatter/classes={
                WIKI={mark=*, draw=cWIKI, fill=cWIKI},
                REDDIT={mark=*, draw=cREDDIT, fill=cREDDIT},
                uci={mark=*, draw=cUCI, fill=cUCI},
                CollegeMsg={mark=*, draw=cCollegeMsg, fill=cCollegeMsg},
                mooc={mark=*, draw=cMOOC, fill=cMOOC},
                LASTFM={mark=*, draw=cLASTFM, fill=cLASTFM},
                Flights={mark=*, draw=cFlights, fill=cFlights},
                GDELT={mark=*, draw=cGDELT, fill=cGDELT},
                SuperUser={mark=*, draw=cSuperUser, fill=cSuperUser}
            },
    ] table [meta=label, col sep=comma] {tikz_data/mrr_2mem.csv};
    \addplot [red, dashed, line width=0.3mm] coordinates {(0,0) (1,1)};
    
    \legend {Wikipedia, REDDIT, UCI, CollegeMsg, MOOC, LASTFM, Flights, GDELT, SuperUser}    
    \end{axis}
\end{tikzpicture}%
    \caption{MRR performance of RNN-based and embedding table-based node memory of all datasets. We fix the neighbor sampling to the most recent strategy and the neighbor aggregator to the attention aggregator. 1-layer models are used with 5-100 neighbors on all datasets. The red diagonal dashed line shows equal performance by RNN-based and embedding table-based node memory.}
    \label{fig:mem}
\end{figure}
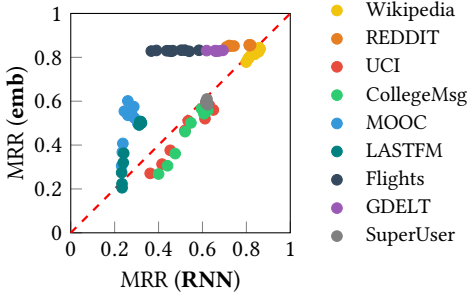

We run 1-layer models with 5 to 100 neighbors and examine the performance of \textbf{emb} and \textbf{RNN} memory on all datasets (see Figure~\ref{fig:mem}). All datasets have results lying on one side of the dividing red line, indicating preferences towards one of the node memory. This preference is not shared across datasets and is independent of other designs.

We explain the preference of datasets towards one type of node memory by closely examining the temporal graph datasets and revealing repetition as a fundamental feature of the datasets. Prior works have analyzed the frequency and effects of edge repetition in some widely used graph datasets~\cite {edgebank, poursafaei2023exhaustive}. While these works show that repetition is common, we further reveal how different repetition patterns affect model performances, especially the performance of node memory. 

Based on a simple intuition, we propose a hypothesis and perform experiments to validate it. We first exemplify our hypothesis with two representative datasets: \textit{Wikipedia} and \textit{Flights}.
\begin{itemize}
    \item \textit{Wikipedia} is an interaction graph of user edits, which exhibits strong temporal locality by automatically saving a user's editing events of the same page repeatedly during an editing session. 
    \item \textit{Flights} records the flights between airports and displays a stable repetition pattern throughout a comparatively long period. 
\end{itemize}

From the perspective of a source node, a recurring event is two interactions with the same destination nodes \newcontent{at different timestamps}. 
For Wikipedia, remembering recent interactions will likely help predict future links because recurring events are temporally dense. For dataset \textit{Flights}, a static vector may suffice to encode all possible interactions as recurring events are periodic.

We introduce \textit{short-term repetition patterns} for describing recurring events with a short time gap in between, and \textit{long-term repetition patterns} for describing recurring events with a long time gap in between. 
With these analyses, our hypothesis is formally stated: Embedding table-based node memory works better on temporal graph datasets with \textit{long-term repetition patterns}, while RNN-based node memory works better on those with \textit{short-term repetition patterns}.

\subsubsection{Measuring dataset repetition.} \label{subsubsec:dsrep}
To reveal the nature of dataset repetition patterns, we show results from three measures: the recurrence matrix, a quantitative indicator \textit{temporal recency ratio} $\phi_\tau$, and distribution fitted to session lengths. 

Recurrence matrices are widely used tools for measuring recurrences of a trajectory $\vec{x}_i \in \mathbb{R}^d$ bounded by an error~\cite{recur}. In our case, $\vec{x}_i\in \mathbb{R}^{|\mathcal{V}|}$ represents an interaction with source node $i$. The time-stamped interactions $\vec{x}_i^{(t)}$ form a trajectory that reflects interactions with source node $i$. In $\vec{x}_i^{(t)}$, indices corresponding to the source and destination nodes are marked 1 and -1, and the other locations are all 0. Error is set to be an arbitrarily small $\epsilon$ so that only interactions with the same destination are considered as recurrences. Algorithm~\ref{alg:recur} shows the construction of recurrence matrix $\mathbf{M}_R$ given bin number $B$. \newcontent{The complexity of this algorithm is $\mathcal{O}(D^2 |\mathcal{V}|)$, where $D$ is the maximum degree. The complexity of this algorithm could be controlled by clipping the maximum degree.}

    

                 

\begin{algorithm}[t]
    \caption{Temporal recurrence matrix generation}  \label{alg:recur}
    \SetKwInOut{KwIn}{Input}
    \SetKwInOut{KwOut}{Output}
    \SetKwInOut{KwSetup}{Setup}
    
    \SetKwFor{For}{for}{do}{end}
    \SetKwFunction{Dst}{Dst}

    \KwIn{ T-CSR graph $\mathcal{G}$, bin number $B$, dataset time range $\left[t_{\min}, t_{\max}\right]$}
    \KwOut{ Recurrence Matrix $\mathbf{M}_R$}
    $\Delta_B \leftarrow \frac{t_{\max}-t_{\min}}{B}$\;
    $\mathbf{M}_{R}\leftarrow \mathbf{0}_{B \times B}$\;
    \newcontent{
    \For{$u \in \mathcal{V}$}{
        \For{$(i,j) \leftarrow (1,2)$ \KwTo $(|\mathcal{N}_{u}|-1,|\mathcal{N}_{u}|)$}{
            $n_i \leftarrow \mathcal{N}_{u}^{(i)}$,
            $n_j \leftarrow \mathcal{N}_{u}^{(j)}$\;
            \uIf{\Dst{$n_i$} $=$ \Dst{$n_j$}}{
                $b_i \leftarrow \frac{t_{n_i}-t_{\min}}{\Delta_B}$,
                $b_j \leftarrow \frac{t_{n_j}-t_{\min}}{\Delta_B}$\;
                $\mathbf{M}_R[b_i][b_j] \leftarrow \mathbf{M}_R[b_i][b_j] + 1 $\;
            }
        }
    }
    }
\end{algorithm}

We run Algorithm~\ref{alg:recur} and show recurrence matrices with heatmaps (See Figure~\ref{fig:recur}). We validate our intuition about the distinct short-term and long-term repetition patterns. \textit{Wikipedia} 
displays bright areas around the diagonal line, showing that recurring events happen within a short period. \textit{REDDIT} 
display vastly different patterns, which indicates that recurring events are highly periodic with long intervals in between. 

We further develop a metric, the temporal recency ratio $\phi_\tau$, to quantitatively assess the positions of datasets along the axis from short- to long-term repetition patterns. This metric is defined based on the $\tau$-recurrence rate~\cite{recur}, denoted as $RR_\tau = \frac{1}{N-\tau}\sum^{B-\tau}_{i=1} \mathbf{R}_{i, i+\tau}$, where $B$ denotes time bins. The $\tau$-recurrence rate calculates the likelihood of an event repeating after $\tau$ time steps.
A distribution of $RR_\tau$ heavily biased towards smaller $\tau$ values indicates short-term repetition patterns, whereas a more uniform distribution on varying $\tau$ signifies long-term repetition patterns.
We define the temporal recency ratio as follows:

\begin{equation}
    \phi_\tau(\mathcal{G}) = \sum_{i=1}^{B} \frac{RR_i}{\sum_{j=1}^{i} RR_j}
\end{equation}

where $B$ is the bin number. $\phi_\tau$ is small if $RR_\tau$ decreases sharply with time (indicative of short-term repetition patterns). It is large if $RR_\tau$ maintains a stable value across different $\tau$ values (indicative of long-term repetition patterns). Additionally, our definition of $\phi_\tau$ brings two advantageous characteristics:

\begin{enumerate}
    \item Independence of time scale: Our datasets have different time scales for their varied sources (Table \ref{tab:ds}). $\phi_\tau$ uniformly measures the repetition regardless of the dilation or shrinkage of time scales.
    \item Independence of repetition frequency: The strength of repetition may vary depending on how events are recorded. $\phi_\tau$ is also independent of repetition strength, focusing solely on distinguishing between long-term and short-term repetition patterns.
\end{enumerate}

We summarize the repetition patterns in the recurrence matrices and \textit{temporal recency ratio} $\phi_\tau (\mathcal{G})$ of our temporal datasets in Table~\ref{tab:recur} \shepherding{and visualize representative datasets in Figure ~\ref{fig:recur}. }\newcontent{There are two noteworthy datasets: 1. \textit{MOOC} dataset is considered an exception by encompassing very sparse repetitions, as shown in Figure ~\ref{fig:recur}. The better performance of \textbf{emb} on the dataset could be attributed to its role in feature augmentation. 2. \textit{SuperUser} is a dataset with both \textit{short-term} and \textit{long-term repetition patterns}, as shown by the recurrence matrix in Figure ~\ref{fig:recur} and the median value of temporal recency ratio. \textbf{emb} and \textbf{RNN} have similar performance on \textit{SuperUser}.}
We found that the recurrence matrices and the temporal recency ratio provide a consistent illustration of dataset repetition patterns, which account for its preference for either dynamic or static node memory. 

\begin{table}[h]
\small
    \setlength{\tabcolsep}{0.9mm}
    \centering
    \caption{Preferred node memory of datasets and the repetition patterns summarized from recurrence matrices and temporal recency ratio $\phi_\tau (\mathcal{G})$.}
    \begin{tabular}{r|ccc}
        \toprule
        Datasets & preferred node memory & repetition patterns & $\phi_\tau (\mathcal{G})$ \\
        \midrule
        \textit{Wikipedia} & \textbf{RNN} & short-term & 0.039 \\
        \textit{UCI} & \textbf{RNN} & short-term & 0.194 \\
        \textit{REDDIT} & \textbf{emb} & long-term & 4.201 \\
        \textit{Flights} & \textbf{emb} & long-term & 4.396 \\
        \textit{LASTFM} & \textbf{emb} & long-term & 3.114 \\
        \newcontent{\textit{MOOC}} & \newcontent{\textbf{emb}} & \newcontent{sparse} & \newcontent{0.333} \\
        \newcontent{\textit{GDELT}} & \newcontent{\textbf{emb}} & \newcontent{long-term} & \newcontent{5.705} \\
        \newcontent{\textit{SuperUser}} & \newcontent{\textbf{RNN}} & \newcontent{mixed} & \newcontent{2.704} \\
        \bottomrule
    \end{tabular}
    \label{tab:recur}
\end{table}

\begin{figure}[h]
    \centering
    \setlength{\abovecaptionskip}{5pt} 
    \setlength{\belowcaptionskip}{-3pt} 
    \includegraphics[width=0.48\textwidth]{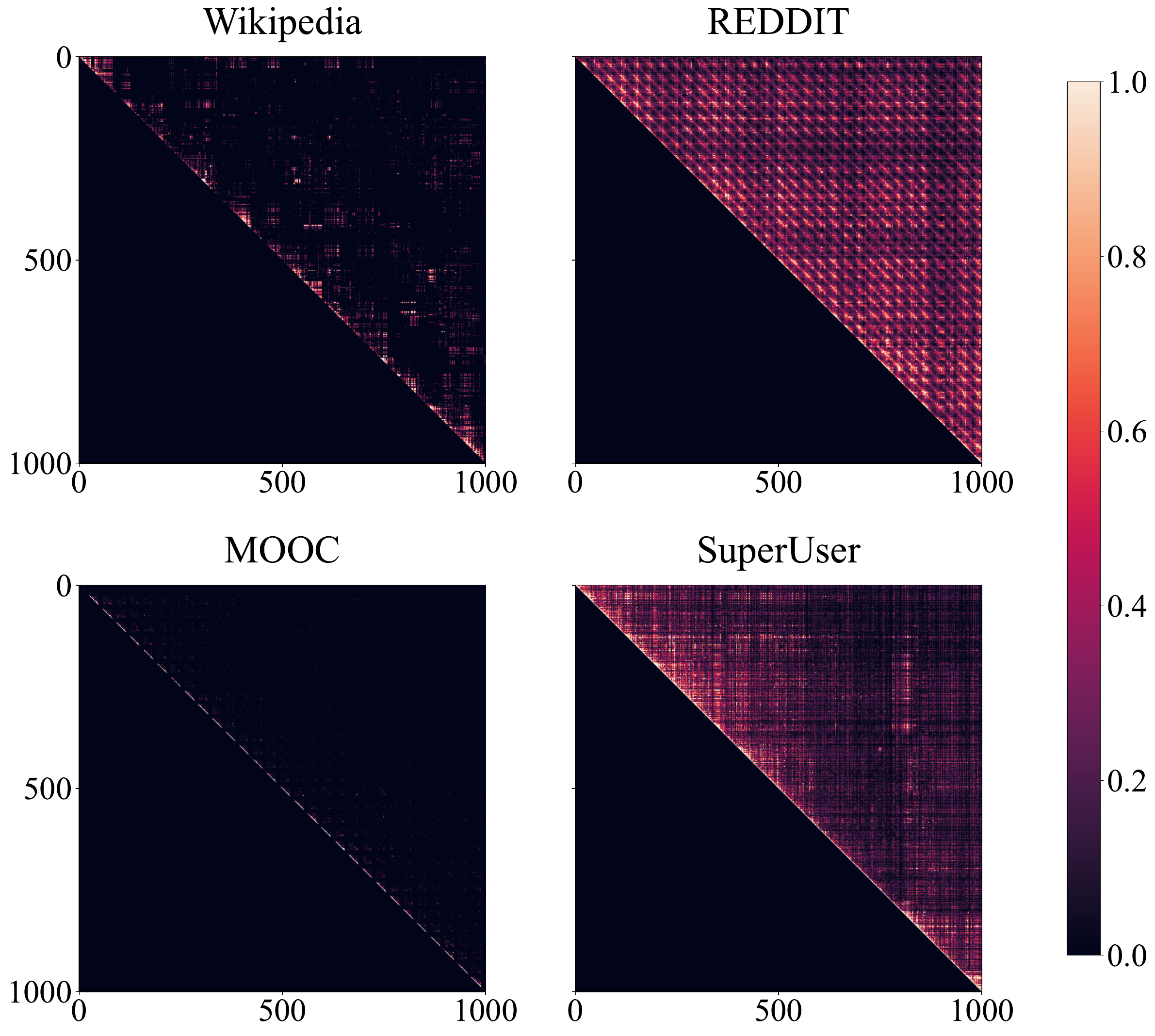}
    \caption{Recurrence matrices of \textit{Wikipedia}, \textit{REDDIT}, \newcontent{\textit{MOOC}, and \textit{SuperUser}}. Extreme values are truncated, and matrices have been normalized. The x-axis and y-axis are bins mapped from timestamps. The value at position $i, j$ in the matrices stands for the count of recurring events at time bins $i$ and $j$. In the heatmap, brighter positions represent larger values, while darker positions represent smaller values. }
    \label{fig:recur}
    
\end{figure}

We also analyze the patterns of session lengths of the datasets and view repetition from the perspective of sessions. We measure periods of consecutive events between the same sources and destinations by $\Delta_{t} = t_{last} - t_{first}$. Following the literature of session identification~\cite{halfaker2015user}, we plot histograms of $\Delta_{t}$ on a logarithmically-scaled x-axis. Session identification methods usually leveraged an inactivity threshold $t_T$ as the estimated cut-off where the probabilities of between-session and within-session reach equality~\cite{halfaker2015user}. We derive that threshold by fitting the histogram under the scaled x-axis to a two-component Gaussian mixture model with a minimal squared difference and finding the intersection point (Figure ~\ref{fig:recur}). Due to space limits, we only display the results from datasets \textit{Wikipedia} and \textit{REDDIT}. 
We find that the datasets with distinguishable long and short sessions (i.e., with a clear intersection point) are those with short-term repetition patterns. In contrast, the datasets with vague or unrecognizable sessions display a long-term periodic repetition or little repetition. This provides a deeper interpretation of \textit{short-term repetition patterns} as sessions and enables us to transfer existing methodology in this area of research to dealing with temporal graph datasets.

\begin{figure}[h]
    \centering
    \setlength{\abovecaptionskip}{5pt} 
    \setlength{\belowcaptionskip}{-3pt} 
    \includegraphics[width=0.45\textwidth]{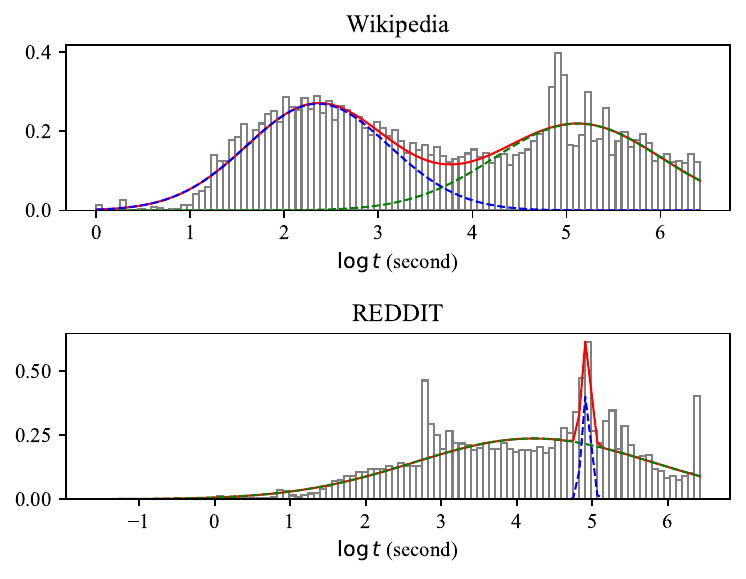}
    \caption{Session length distributions of datasets. The x-axis represents time on a logarithmic scale, and the y-axis represents the probability distribution density. The blue and green dashed lines represent two Gaussian distributions, while the red line represents the mixed distribution. }
    \label{fig:recur}
    
\end{figure}

Finally, we summarize our analyses and restate our claim. With the measurement of dataset repetition using both recurrence matrices and temporal recency ratio, we categorize the dataset repetition into \textit{short-term} and \textit{long-term} accordingly. We observe an alignment between the repetitions and a preference towards \textbf{RNN} or \textbf{emb} node memory: datasets with \textit{long-term repetition patterns} have a strong preference towards \textbf{emb} node memory, while those with \textit{short-term repetition patterns} tend to favor \textbf{RNN}. 

\subsubsection{Validation on synthetic datasets.}
To further minimize interferences and substantiate our claim, we generate synthetic datasets with a more controllable experimental setting than often noisy and biased real-world datasets. We construct the dataset by sequentially generating events at every time step. First, we sample the timestamp of the events by adding a small deviation to the current time step. Second, the source node is sampled from a predefined distribution $f$. Third, the destination node is sampled either from historical events or the entire node set. The process for generating a synthetic temporal graph dataset is described in Algorithm ~\ref{alg:gen-syn}.

    


\begin{algorithm}[t]
    \caption{Synthetic temporal graph generation} \label{alg:gen-syn}
    \SetKwInOut{KwIn}{Input}
    \SetKwInOut{KwOut}{Output}
    \SetKwInOut{KwSetup}{Setup}
    
    \SetKwFor{For}{for}{do}{end}
    \SetKwFunction{Dst}{Dst}

    \KwIn{node count $N$, edge count $E$, timesteps $T$, cluster coefficient $C_i \sim \mathtt{N}(\mu, \sigma^2)$,  long/short-term controller $\alpha \in (0, 1]$, repetition strength controller $\beta \in (0, 1]$, temperature $\mathbf{T}$}
    \KwOut{$\mathcal{G} = (\mathcal{V},\mathcal{E})$ with events $\{(i, j, \bm {e}_{ijt}, t_{\bm e})\}$}
    \newcontent{
        \For{$(t,k) \leftarrow (1,1)$ \KwTo $(T, \frac{E}{T})$}{
            $\delta_{\bm e } \leftarrow \delta_e \sim \mathtt{N}(0, \sigma_t^2)$ \;
            $t_{\bm e }= t + \delta_{\bm e }$ \;
            Sample source node $i$ with $p(i) \propto C_i$ \;
            \uIf{$\mathcal{N}(i, t)=\emptyset$}{
                Sample destination node $j$ with $p(j) \propto C_i$ \;
            }
            \Else{
                With probability $\beta$, sample destination node $j$ from $\{dst(\bm e), \bm e\in \mathcal{N}(i, t)\}$ with $p(j) \propto \alpha^{\frac{t-t_{\bm e }}{\mathbf{T}}}$ \;
                With probability $1 - \beta$, sample destination node $j$ with $p(j) \propto C_i$ \;
            }
        }
    }
\end{algorithm}

\begin{figure}[tb]
    \centering
    \setlength{\abovecaptionskip}{5pt} 
    \setlength{\belowcaptionskip}{-3pt} 
    \includegraphics[width=0.48\textwidth]{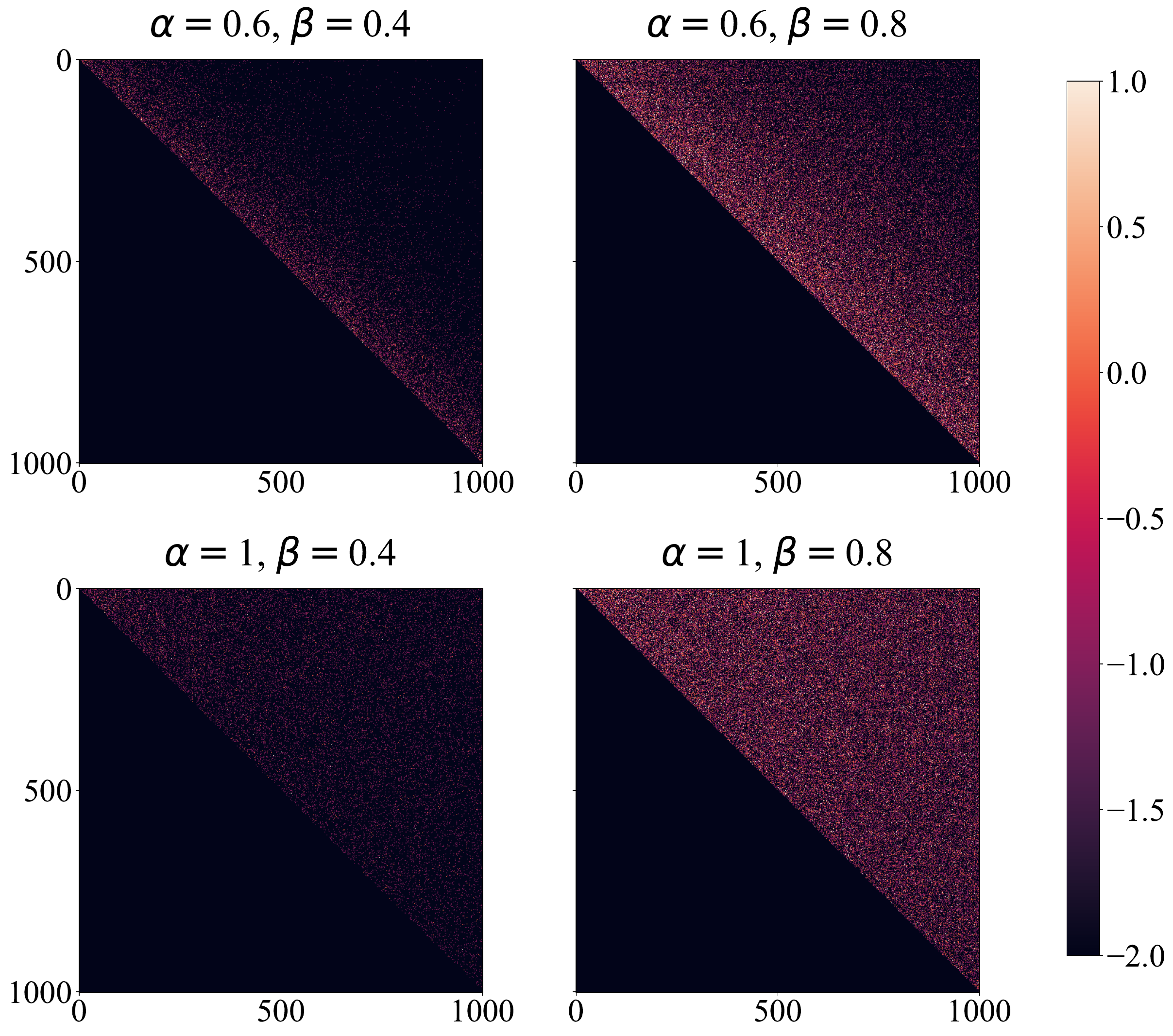}
    \caption{Recurrence matrices of synthetic datasets with different $\alpha$ and $\beta$. In every column, larger $\alpha$ corresponds to longer-term repetition patterns. In every row, larger $\beta$ corresponds to more repetitions.}
    \label{fig:syn-recur}
\end{figure}

Using $\alpha$, we construct an exponential-decaying probability distribution of sampling historical neighbors. $\alpha$ approaching 1 represents uniformly sampling all historical neighbors, while $\alpha$ approaching 0 represents only sampling the most recent temporal neighbor. The bias of $\alpha$ towards these two extremes corresponds to \textit{long-term} and \textit{short-term} repetition, respectively. We use $\beta$ to control the intensity of repetition, where a larger $\beta$ represents stronger repetition.
We show recurrence matrices of the synthetic datasets with different combinations of $\alpha$ and $\beta$ in Figure ~\ref{fig:syn-recur}.

\begin{figure}[b]
    \centering
    \setlength{\abovecaptionskip}{5pt} 
    \setlength{\belowcaptionskip}{-3pt} 
    \tikzsetnextfilename{mrr_syn}
\begin{tikzpicture}
    
    \definecolor{c1}{RGB}{231, 76, 60}
    \definecolor{c2}{RGB}{230, 126, 34}
    \definecolor{c3}{RGB}{241, 196, 15}
    \definecolor{c4}{RGB}{46, 204, 113}
    \definecolor{c5}{RGB}{52, 152, 219}
    \definecolor{c6}{RGB}{0, 128, 128}
    \definecolor{c7}{RGB}{52, 73, 94}
    \definecolor{c8}{RGB}{0, 0, 128}
    \definecolor{c9}{RGB}{230, 101, 47}
    \definecolor{c10}{RGB}{236, 161, 24}
    \definecolor{c11}{RGB}{144, 200, 64}
    \definecolor{c12}{RGB}{49, 178, 166}
    \definecolor{c13}{RGB}{26, 140, 174}
    \definecolor{c14}{RGB}{26, 100, 111}
    \definecolor{c15}{RGB}{26, 36, 111}

    \begin{groupplot}[
        group style={
            group size=1 by 1,
            horizontal sep=0.9cm,
            vertical sep=0.4cm
        },
        tick label style={font=\scriptsize},
        label style={font=\small},
        width=5cm,
        height=4cm,
    ]
    
    \nextgroupplot[
    xlabel={$\alpha$}, xlabel style={yshift=2ex},
    ylabel={MRR},ylabel style={yshift=-4ex},grid=major,
        ymin=0.10,ymax=0.32,xmin=0.8,xmax=4.2,xticklabel style={inner sep=2pt}, xtick pos=lower,
        scaled x ticks=false, xtick={1,2,3,4}, xticklabels={0.6,0.7,0.8,0.9}, ytick={0.10, 0.15,0.20,0.25,0.30,0.35,0.40},
        legend pos=south east,legend style={legend columns=1,font=\fontsize{7}{8}\selectfont},legend image post style={xscale=0.75},
        ]

    \addplot[color=c1,line width=0.25mm, error bars/.cd,  y fixed, y dir=both, y explicit,] table[x expr=\coordindex+1, y=mrr_gru_0.4, y error=std_gru_0.4,col sep=comma]{tikz_data/mrr_syn.csv}; \label{f9l11}
    \addplot[color=c3,line width=0.25mm, error bars/.cd,  y fixed, y dir=both, y explicit,] table[x expr=\coordindex+1, y=mrr_gru_0.5, y error=std_gru_0.5,col sep=comma]{tikz_data/mrr_syn.csv}; \label{f9l12}
    \addplot[color=c4,line width=0.25mm, error bars/.cd,  y fixed, y dir=both, y explicit,] table[x expr=\coordindex+1, y=mrr_gru_0.6, y error=std_gru_0.6,col sep=comma]{tikz_data/mrr_syn.csv}; \label{f9l13}
        
    \addplot[color=c1,line width=0.25mm, dashed, dash pattern=on 1pt off 0.5pt,  error bars/.cd,  y fixed, y dir=both, y explicit,] table[x expr=\coordindex+1, y=mrr_emb_0.4, y error=std_emb_0.4,col sep=comma]{tikz_data/mrr_syn.csv}; \label{f9l21}
    \addplot[color=c3,line width=0.25mm, dashed, dash pattern=on 1pt off 0.5pt,  error bars/.cd,  y fixed, y dir=both, y explicit,] table[x expr=\coordindex+1, y=mrr_emb_0.5, y error=std_emb_0.5,col sep=comma]{tikz_data/mrr_syn.csv}; \label{f9l22}
    \addplot[color=c4,line width=0.25mm, dashed, dash pattern=on 1pt off 0.5pt,  error bars/.cd,  y fixed, y dir=both, y explicit,] table[x expr=\coordindex+1, y=mrr_emb_0.6, y error=std_emb_0.6,col sep=comma]{tikz_data/mrr_syn.csv}; \label{f9l23}

    \end{groupplot}
    
    \node[right] at ($(group c1r1.east)-(-3pt,0.0cm)$) {
    \begin{tikzpicture}
        \matrix[matrix of nodes, inner sep=0.2em, draw=none, ]{
                \ref{f9l11} & RNN, $\beta=0.4$   & [5pt]   \\
                \ref{f9l12} & RNN, $\beta=0.5$ & [5pt]   \\
                \ref{f9l13} & RNN, $\beta=0.6$ & [5pt]   \\
                \ref{f9l21} & emb, $\beta=0.4$  & [5pt]   \\
                \ref{f9l22} & emb, $\beta=0.5$   & [5pt]   \\
                \ref{f9l23} & emb, $\beta=0.6$  & [5pt]    \\ 
            };
    \end{tikzpicture}
    };
\end{tikzpicture}%
    \caption{MRRs of TGNN models with \textbf{emb} and \textbf{RNN} node memory, with respect to $\alpha$ and $\beta$. We use a 1-layer model with 10 neighbors, \textbf{MR} neighbor sampler, and \textbf{atten} neighbor aggregator.}
    \label{fig:syn-mem}
\end{figure}
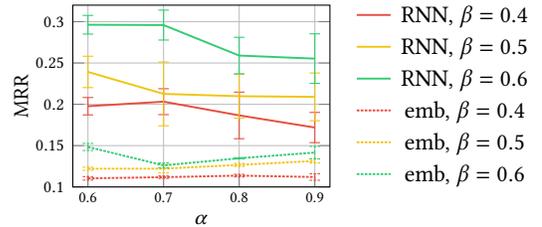

Using synthetic datasets, we conduct controllable experiments on how dataset repetition patterns affect different node memory performance. Our results are summarized in Figure ~\ref{fig:syn-mem}. We observe that with a larger $\alpha$ (corresponding to long-term repetitive pattern), \textbf{RNN} node memory tends to have a decreased performance, while \textbf{emb} node memory performance gradually increases, supporting our previous hypothesis on dataset repetition affecting node memory performance.

\subsubsection{Discussions in the inductive setting}

We note that \textbf{emb} node memory has a drawback in the inductive experimental setting. As the node memory is only updated during the training stage, no information from validation or test can be integrated into node memory to assist future tasks. The new nodes appearing only in the test stage would have only randomized vectors as their node embeddings. Our experiments show a significant drop in test accuracy compared to the transductive setting.

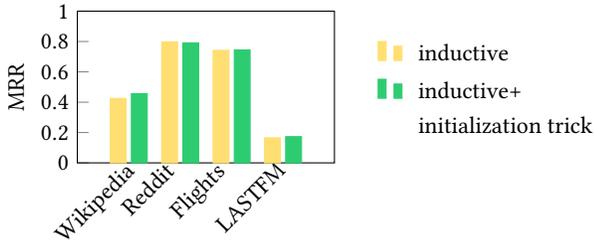
\begin{figure}[t]
    \centering
    \setlength{\abovecaptionskip}{5pt} 
    \setlength{\belowcaptionskip}{-3pt} 
    \tikzsetnextfilename{embed_strategy}
\begin{tikzpicture}

    \definecolor{c2}{RGB}{46, 204, 113}
    \definecolor{c1}{RGB}{255, 223, 113}
    
    \begin{axis}[
        ybar, 
        xtick={1,2,3,4},
        xticklabels={Wikipedia,Reddit,Flights,LASTFM}, 
        major x tick style=transparent,
        xtick pos=lower,
        xticklabel style={rotate=45,xshift=-0.5ex,yshift=4ex},
        ylabel={MRR}, ylabel near ticks, ylabel style={yshift=0ex},
        ytick pos=left,
        bar width=6pt,
        ymin=0, ymax=1, 
        xmin=0, xmax=5,
        width=5cm,
        height=3.6cm,
        name=ax1,
    ]
        \addplot[style={c1,fill=c1,mark=none}] table[col sep=comma, x expr=\coordindex+1, y=before]{tikz_data/embed_strategy.csv}; \label{bl1}
        \addplot[style={c2,fill=c2,mark=none}] table[col sep=comma, x expr=\coordindex+1, y=after]{tikz_data/embed_strategy.csv}; \label{bl2}
        
    \end{axis}

    \matrix (legend) at ([xshift=2cm] ax1.east) [matrix of nodes, nodes={anchor=west}] {
        \ref{bl1} & inductive \\
        \ref{bl2} & inductive+ \\
        &initialization trick \\
    };

\end{tikzpicture}%
    \caption{MRRs of models with embedding table-based node memory tested on inductive settings. We fix models to be 1-layer with 10 neighbors, using attention neighbor aggregators and the most recent neighbor sampling. MRRs of models w/ and w/o the embedding initialization trick for inductive settings are displayed.}
    \label{fig:emb-strategy}
\end{figure}

Given the strength of \textbf{emb} node memory shown on multiple datasets, we propose a preliminary measure to alleviate the drawbacks of \textbf{emb} node memory. We leverage the mailbox in \textbf{RNN} node memory to store the messages when events are observed for inductive nodes. We use the node memory of 1-hop or 2-hop neighbors as messages, depending on whether the graph is bipartite. When the mailbox has enough mails (in our experiments, we set $\# mails = 3$), we smooth the messages as the initial memory of the inductive node. This initialization trick improves inductive MRRs by $\sim$ 3\% at best (Figure ~\ref{fig:emb-strategy}). Deliberately designed strategies that combine static and dynamic node embeddings in transductive and inductive settings will likely yield better performance. It should be mentioned that online fine-tuning~\cite{gnnflow} with streaming batches is also applicable in the continuous learning setting.

\subsection{Connections and choices between modules}\label{subsec:conn}

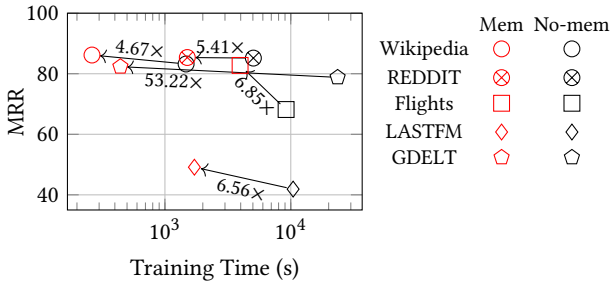
\begin{figure}[h]
  \centering
  \setlength{\abovecaptionskip}{5pt} 
  \setlength{\belowcaptionskip}{-3pt} 
  \definecolor{ours}{RGB}{255,0,0}
\definecolor{DySAT}{RGB}{0,0,0}
\definecolor{JODIE}{RGB}{0,0,0}
\definecolor{TGAT}{RGB}{0,0,0}
\definecolor{TGN}{RGB}{0,0,0}
\definecolor{APAN}{RGB}{0,0,0}

\tikzsetnextfilename{memory_time}
\begin{tikzpicture}
    \begin{axis}[xmode=log,width=5.5cm,height=4.2cm,xlabel={Training Time (s)},ylabel={MRR},ymin=35,ymax=100,grid=major,ylabel style={yshift=-4ex}]
        
        \addplot[mark options={opacity=0},mark=o,->,mark size=3,shorten >=3,shorten <=3] coordinates{(1473.40, 83.22) (263.33, 86.18)};
        \addplot[mark options={color=ours},color=white,mark=o,mark size=3]coordinates{(263.33, 86.18)};
            \label{mwiki}
        \addplot[mark options={color=JODIE},color=white,mark=o,mark size=3]coordinates{(1473.40, 83.22)};
            \label{nmwiki}
        \path (axis cs:263.33, 86.18) -- (axis cs:1473.40, 83.22) node [midway, above=-0.5mm, sloped] {\small4.67$\times$};

        \addplot[mark options={opacity=0},mark=otimes,->,mark size=3,shorten >=3,shorten <=3] coordinates{(5044.42, 85.17) (1505.96, 85.36)};
        \addplot[mark options={color=ours},color=white,mark=otimes,mark size=3]coordinates{(1505.96, 85.36)};
            \label{mreddit}
        \addplot[mark options={color=TGAT},color=white,mark=otimes,mark size=3]coordinates{(5044.42, 85.17)};
            \label{nmreddit}
        \path (axis cs:5044.42, 85.17) -- (axis cs:1505.96, 85.36) node [midway, above=-0.5mm, sloped] {\small5.41$\times$};

        \addplot[mark options={opacity=0},mark=square,->,mark size=3,shorten >=3,shorten <=3] coordinates{(9181.15, 68.18) (3957.15, 82.75)};
        \addplot[mark options={color=ours},color=white,mark=square,mark size=3]coordinates{(3957.15, 82.75)};
            \label{mfli}
        \addplot[mark options={color=TGN},color=white,mark=square,mark size=3]coordinates{(9181.15, 68.18)};
            \label{nmfli}
        \path (axis cs:3957.15, 82.75) -- (axis cs:9181.15, 68.18) node [midway, below=-0.5mm, sloped] {\small6.85$\times$};

        \addplot[mark options={opacity=0},mark=diamond,->,mark size=3,shorten >=2.5,shorten <=2.5] coordinates{(10439.53, 41.89) (1714.38, 49.13)};
        \addplot[mark options={color=ours},color=white,mark=diamond,mark size=3]coordinates{(1714.38, 49.13)};
            \label{mlast}
        \addplot[mark options={color=APAN},color=white,mark=diamond,mark size=3]coordinates{(10439.53, 41.89)};
            \label{nmlast}
        \path (axis cs:1714.38, 49.13) -- (axis cs:10439.53, 41.89) node [midway, below=-0.5mm, sloped] {\small6.56$\times$};

        \addplot[mark options={opacity=0},mark=pentagon,->,mark size=3,shorten >=2.5,shorten <=2.5] coordinates{ (23562.14, 78.80) (442.77, 82.34)};
        \addplot[mark options={color=ours},color=white,mark=pentagon,mark size=3]coordinates{(442.77, 82.34)};
            \label{mgdelt}
        \addplot[mark options={color=APAN},color=white,mark=pentagon,mark size=3]coordinates{(23562.14, 78.80)};
            \label{nmgdelt}
        \path (axis cs:23562.14, 78.80) -- (axis cs:442.77, 82.34) node [pos=0.75, below=-0.5mm, sloped] {\small53.22$\times$};

        
        \coordinate (right) at (rel axis cs:1,0);

    \end{axis}    

    \matrix[matrix of nodes,anchor=north west,inner sep=0.2em,draw=none,font=\small,column 5/.style={anchor=base west}] at ([xshift=0.1cm,yshift=2.7cm] right)
    {
         & Mem & No-mem\\
        Wikipedia & \ref{mwiki} & \ref{nmwiki}\\
        REDDIT & \ref{mreddit} & \ref{nmreddit}\\
        Flights & \ref{mfli} & \ref{nmfli}\\
        LASTFM & \ref{mlast} & \ref{nmlast}\\
        GDELT & \ref{mgdelt} & \ref{nmgdelt}\\
    };
\end{tikzpicture}%
  \caption{\newcontent{Total training time and MRRs of models w/ and w/o node memory. We use \textbf{MR} neighbor sampler and \textbf{atten} aggregator for both models. We compare the 1-layer model with 5 neighbors w/ memory to the 2-layer model 40 supporting neighbors per layer w/o node memory. On each dataset, we select the better-performing node memory between \textbf{emb} and \textbf{RNN}.}}
  \label{fig:mem_time}
\end{figure}

After introducing the neighbor sampling and node memory modules, we explore strategies for effective combined utilization and emphasize the importance of incorporating both modules. Specifically, we evaluate the cost-effectiveness of incorporating node memory compared to models that are only based on neighbor aggregation(e.g., TGAT~\cite{tgat}).
We compare the cost-effectiveness between two approaches: 1. Adopting a deeper model and increasing the size of supporting neighbors; 2. \newcontent{Maintaining node memory in a shallow model with a small supporting neighborhood. }
\newcontent{Our results in Figure ~\ref{fig:mem_time} demonstrate that models with node memory generally produce better results than pure sampling with less computation cost. }

\subsection{Discussions on other modules} \label{subsec:others}

Finally, we provide clear conclusions on neighbor aggregators and time encodings. While this section includes limited discussion, our results highlight equally meaningful factors to consider in TGNN designs as the previous sections.

\subsubsection{Neighbor aggregators} \label{subsubsec:aggr}
\begin{figure}[h]
    \centering
    \setlength{\abovecaptionskip}{5pt} 
    \setlength{\belowcaptionskip}{-3pt} 
    \input{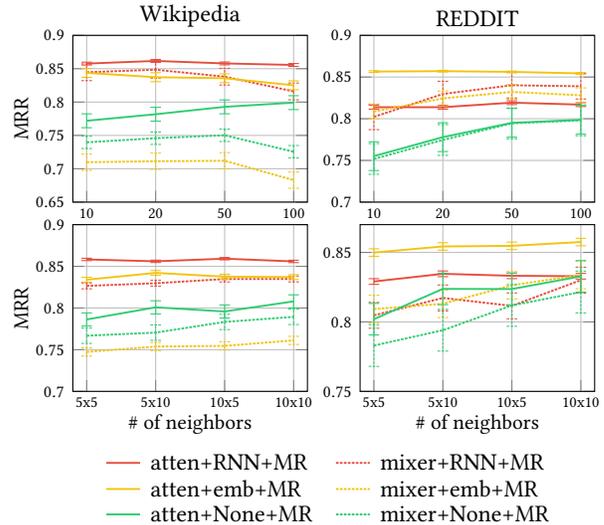}
    \caption{MRRs and runtimes of models with an increasing size of sampled neighbors on datasets \textit{Wikipedia} and \textit{REDDIT}. We show results of \textbf{atten} and \textbf{mixer} neighbor aggregators using solid and dashed lines under various settings on 1-layer (first row) and 2-layer (second row) models.}
    \label{fig:aggr}
    
\end{figure}

Despite that \textbf{mixer} is a method proposed more recently, we found that \textbf{atten} neighbor aggregator generally outperforms \textbf{mixer} in a variety of settings (See Figure~\ref{fig:aggr}). \shepherding{This phenomenon is universally observed on all datasets.} We observe that the solid lines standing for \textbf{atten} aggregators usually outperform their dashed-line counterparts standing for \textbf{mixer} aggregators by a large margin. While the superiority of \textbf{atten} neighbor aggregator is only displayed on two datasets due to space limitations, this phenomenon has been observed on all experimented datasets. \shepherding{Despite the theoretical quadratic complexity of \textbf{atten}, \textbf{atten} has a shorter runtime ($<50\%$ when training in large scale) than \textbf{mixer} for our sequence length is usually limited.}

\subsubsection{Time encoding}
Most existing works~\cite{tgn, tgat, apan} have employed a learnable time encoding $\boldsymbol{\Phi}$:
\begin{equation}
    \boldsymbol{\Phi} \left( \Delta t \right) = \text{cos}\left( \Delta t \boldsymbol{w} + \boldsymbol{b} \right),
\end{equation}
where $\boldsymbol{w} \in \mathbb{R}^{d_T}$ and $\boldsymbol{b} \in \mathbb{R}^{d_T}$ are learnable parameters.

GraphMixer~\cite{graphmixer} proposed a non-learnable time encoding for events and claims the encoder exhibits smoother optimization, faster convergence, and a better generalization:
\begin{equation} \label{eq:time-encode}
    \boldsymbol{\Phi} \left( \Delta t \right) = \text{cos}\left( \Delta t \boldsymbol{\omega} \right),\quad\boldsymbol{\omega}=\left\{\alpha^{-(i-1) / \beta}\right\}_{i=1}^{d_T},
\end{equation}

\shepherding{Our experiment results align with the claim of ~\cite{graphmixer} that non-learnable time encoding outperforms learnable time encoding. }

\section{Conclusion}\label{sec:discuss}
In this work, we proposed a modular TGNN comparison framework with reasonable optimizations, addressing the problem of insufficient design space search and unfair comparison. 
\newcontent{While neighbor sampling and aggregation are problems discussed in both temporal and static graphs, we focus on temporal graphs to reveal domain knowledge about the TGNN modules.}
Through extensive experiments, we drew conclusions on the cost-effectiveness of TGNN modules with different sizes and choices, the dependence between module preferences and datasets, and the relations between different modules. Based on our findings, we now revisit the three research questions presented in Section~\ref{sec:intro}.

\noindent\textit{RQ1. Efficiency and Cost-effectiveness of Module Designs: What designs strike the optimal balance between effectiveness and resource efficiency? }
We demonstrate that {\bf TGNNs should adopt most recent neighbor sampling, attention-based neighbor aggregator, and maintain a node memory for effectiveness and efficiency}.
We showed that: 1. With a similar or lower runtime, most recent neighbor sampling consistently outperforms uniform neighbor sampling. 2. Despite the fact that MLP-Mixer is proposed later than attention aggregator, attention neighbor aggregator outperforms the MLP-Mixer neighbor aggregator. 3. With comparable or even better model performance, we demonstrated that maintaining node memory is more efficient than increasing the number and layer of neighbor sampling to achieve comparable results. 

\noindent\textit{RQ2. Universality of Module Effectiveness across Datasets:
Do the best-performing modules on some datasets maintain their effectiveness universally across all datasets, or is module performance dependent on dataset characteristics?}
For neighbor sampling, neighbor aggregator, and time encoding, our experiments show consistent results of accuracy and runtime comparisons across all datasets. For node memory, our results show that {\bf datasets exhibit a preference for different types of node memory based on whether their repetition patterns are long-term or short-term}. 
\shepherding{We demonstrate that the former kind of dataset prefers embedding table-based node memory, while the latter prefers RNN-based node memory.}

\noindent\textit{RQ3. Interplay between Different Modules in the Model: Does the integration of various modules enhance or undermine performance?}
We observe that {\bf when applying node memory, a deeper or wider neighbor sampling often brings little performance gain} to the model. This can be attributed to the overlapping effect of an enlarged receptive field brought by using node memory and sampling a large temporal neighborhood. We also find that combining RNN-based node memory and uniform neighbor samplers yields a poor performance for potentially not using the latest information to update the RNN-based node memory.


\section{Related work} \label{sec:related}


\noindent\textbf{Benchmarking temporal graph learning.}
Recent Temporal Graph Learning benchmark works provide platforms for comparisons and conducting empirical comparisons of models.

EdgeBank~\cite{edgebank} illustrates the limitations of tasks and metrics by showing the high prediction accuracy of a simple baseline. They also analyze dataset deficiency and enhance evaluation.
TGB~\cite{tgb}, BenchTemp~\cite{benchtemp}, and DyGLib~\cite{dyglib} provide evaluation pipelines for temporal graph learning. 
Their comparisons found that the performance of models drastically varies on tasks and datasets. 

On the other hand, some works focus on conducting empirical model comparisons and drawing meaningful conclusions from results. \newcontent{Chen et al.~\cite{chen2022bottleneck}} conduct a detailed bottleneck profiling of the runtime of TGNN models. \newcontent{Gravina et al.~\cite{gravina2024deep}} conduct experiments on node- and edge-level tasks using recent TGNNs. 

However, existing works lack the support for design space exploration. Moreover, benchmarking is performed using different model implementations, often without a reasonable amount of optimization. In this work, we made extensive efforts to fill this gap.

\noindent \textbf{Improving the efficiency of temporal graph neural networks.}
A variety of works have laid emphasis on increasing the efficiency of TGNNs.
TGL~\cite{tgl} focuses on resolving scalability challenges by exploring parallelizing opportunities using time-centric data structures, mechanisms, and graph engines. \newcontent{Gangda et al.~\cite{taser}} introduce an adaptive temporal graph sampling algorithm to achieve better results with smaller sampling sizes. \newcontent{Alomrani et al.~\cite{alomrani2023dyg2vec}} leverage temporal edge encodings and window-based subgraph sampling to generate embeddings.

\newcontent{On top of the TGNN model, various methods accelerate TGNNs by leveraging the characteristics of some TGNNs. DistTGL~\cite{disttgl} and GNNFlow~\cite{gnnflow} are two distributed frameworks designed for faster training paradigms and comparable performance. }
GNNFlow ~\cite{gnnflow} also presents support for continuous TGNN training. \newcontent{TGOpt~\cite{tgopt}} reduces the redundancies to accelerate the inference performance of TGAT~\cite{tgat}. \newcontent{Orca~\cite{orca}} caches intermediate embeddings for reusing to reduce redundant computations.
\newcontent{Shihong et al.~\cite{etc}} accelerate TGNN training by improving the data batching and accessing manner to enlarge the batch size and reduce access volume.

\section{Future directions} \label{sec:future}


\noindent\textbf{Exploring additional benchmarking aspects.} Future works can extend the benchmarking efforts of our work, such as exploring the effects of i) sampling with exponential decaying probabilities, ii) combining static and dynamic node memories, and iii) using different methods for negative sample selection. Comparison and validation of modules may also be carried out on random walk-based TGNN models. In addition to prediction accuracy, evaluation can be expanded to model robustness, generalization, and privacy.

\noindent\textbf{Improving the design of TGNN modules.} Existing modules exhibit limited adaptability on datasets. For example, one node memory type only performs well on short- or long-term repetitive datasets. \newcontent{Given that valuable information is distributed in various forms over past time periods, it is important to explore how more universal TGNN modules can be developed to effectively capture this information across diverse datasets and learning settings. }

\noindent\textbf{Examining dataset patterns.} We demonstrated that TGNN dataset patterns remain largely unexplored, potentially preventing us from designing useful and efficient neighbor aggregators and neighbor sampling methods. Besides dataset repetition patterns discussed in this work, dataset heterophily and frequency domain patterns may be investigated.

\noindent\textbf{Temporal neighbor modeling and sampling methods.} Our findings indicated an orderly distribution of historical neighbors that could be fitted to some probability distributions. This insight can enhance temporal neighbor modeling or enable smart sampling that aggregates more informative neighbors with a lower budget. 



\bibliographystyle{ACM-Reference-Format}
\bibliography{ref}


\begin{thebibliography}{60}


\ifx \showCODEN    \undefined \def \showCODEN     #1{\unskip}     \fi
\ifx \showDOI      \undefined \def \showDOI       #1{#1}\fi
\ifx \showISBNx    \undefined \def \showISBNx     #1{\unskip}     \fi
\ifx \showISBNxiii \undefined \def \showISBNxiii  #1{\unskip}     \fi
\ifx \showISSN     \undefined \def \showISSN      #1{\unskip}     \fi
\ifx \showLCCN     \undefined \def \showLCCN      #1{\unskip}     \fi
\ifx \shownote     \undefined \def \shownote      #1{#1}          \fi
\ifx \showarticletitle \undefined \def \showarticletitle #1{#1}   \fi
\ifx \showURL      \undefined \def \showURL       {\relax}        \fi
\providecommand\bibfield[2]{#2}
\providecommand\bibinfo[2]{#2}
\providecommand\natexlab[1]{#1}
\providecommand\showeprint[2][]{arXiv:#2}

\bibitem[\protect\citeauthoryear{Alomrani, Biparva, Zhang, and Coates}{Alomrani et~al\mbox{.}}{2023}]%
        {alomrani2023dyg2vec}
\bibfield{author}{\bibinfo{person}{Mohammad Alomrani}, \bibinfo{person}{Mahdi Biparva}, \bibinfo{person}{Yingxue Zhang}, {and} \bibinfo{person}{Mark Coates}.} \bibinfo{year}{2023}\natexlab{}.
\newblock \showarticletitle{DyG2Vec: Efficient Representation Learning for Dynamic Graphs}.
\newblock \bibinfo{journal}{\emph{Transactions on Machine Learning Research}} (\bibinfo{year}{2023}).
\newblock


\bibitem[\protect\citeauthoryear{Bai, Zhang, Wu, and Nie}{Bai et~al\mbox{.}}{2020}]%
        {bai2020temporal}
\bibfield{author}{\bibinfo{person}{Ting Bai}, \bibinfo{person}{Youjie Zhang}, \bibinfo{person}{Bin Wu}, {and} \bibinfo{person}{Jian-Yun Nie}.} \bibinfo{year}{2020}\natexlab{}.
\newblock \showarticletitle{Temporal graph neural networks for social recommendation}. In \bibinfo{booktitle}{\emph{2020 IEEE International Conference on Big Data (Big Data)}}. IEEE, \bibinfo{pages}{898--903}.
\newblock


\bibitem[\protect\citeauthoryear{Behrouz, Hashemi, Sadeghian, and Seltzer}{Behrouz et~al\mbox{.}}{2024}]%
        {catwalk}
\bibfield{author}{\bibinfo{person}{Ali Behrouz}, \bibinfo{person}{Farnoosh Hashemi}, \bibinfo{person}{Sadaf Sadeghian}, {and} \bibinfo{person}{Margo Seltzer}.} \bibinfo{year}{2024}\natexlab{}.
\newblock \showarticletitle{CAT-Walk: Inductive Hypergraph Learning via Set Walks}.
\newblock \bibinfo{journal}{\emph{Advances in Neural Information Processing Systems}}  \bibinfo{volume}{36} (\bibinfo{year}{2024}).
\newblock


\bibitem[\protect\citeauthoryear{Bui, Cho, and Yi}{Bui et~al\mbox{.}}{2022}]%
        {bui2022spatial}
\bibfield{author}{\bibinfo{person}{Khac-Hoai~Nam Bui}, \bibinfo{person}{Jiho Cho}, {and} \bibinfo{person}{Hongsuk Yi}.} \bibinfo{year}{2022}\natexlab{}.
\newblock \showarticletitle{Spatial-temporal graph neural network for traffic forecasting: An overview and open research issues}.
\newblock \bibinfo{journal}{\emph{Applied Intelligence}} \bibinfo{volume}{52}, \bibinfo{number}{3} (\bibinfo{year}{2022}), \bibinfo{pages}{2763--2774}.
\newblock


\bibitem[\protect\citeauthoryear{Cai, Chen, Luo, Gui, Ni, Li, and Chen}{Cai et~al\mbox{.}}{2021}]%
        {cai2021structural}
\bibfield{author}{\bibinfo{person}{Lei Cai}, \bibinfo{person}{Zhengzhang Chen}, \bibinfo{person}{Chen Luo}, \bibinfo{person}{Jiaping Gui}, \bibinfo{person}{Jingchao Ni}, \bibinfo{person}{Ding Li}, {and} \bibinfo{person}{Haifeng Chen}.} \bibinfo{year}{2021}\natexlab{}.
\newblock \showarticletitle{Structural temporal graph neural networks for anomaly detection in dynamic graphs}. In \bibinfo{booktitle}{\emph{Proceedings of the 30th ACM international conference on Information \& Knowledge Management}}. \bibinfo{pages}{3747--3756}.
\newblock


\bibitem[\protect\citeauthoryear{Cao, Wang, Duan, Zhang, Zhu, Huang, Tong, Xu, Bai, Tong, et~al\mbox{.}}{Cao et~al\mbox{.}}{2020}]%
        {cao2020spectral}
\bibfield{author}{\bibinfo{person}{Defu Cao}, \bibinfo{person}{Yujing Wang}, \bibinfo{person}{Juanyong Duan}, \bibinfo{person}{Ce Zhang}, \bibinfo{person}{Xia Zhu}, \bibinfo{person}{Congrui Huang}, \bibinfo{person}{Yunhai Tong}, \bibinfo{person}{Bixiong Xu}, \bibinfo{person}{Jing Bai}, \bibinfo{person}{Jie Tong}, {et~al\mbox{.}}} \bibinfo{year}{2020}\natexlab{}.
\newblock \showarticletitle{Spectral temporal graph neural network for multivariate time-series forecasting}.
\newblock \bibinfo{journal}{\emph{Advances in neural information processing systems}}  \bibinfo{volume}{33} (\bibinfo{year}{2020}), \bibinfo{pages}{17766--17778}.
\newblock


\bibitem[\protect\citeauthoryear{Chang, Gao, Zheng, Hui, Niu, Song, Jin, and Li}{Chang et~al\mbox{.}}{2021}]%
        {chang2021sequential}
\bibfield{author}{\bibinfo{person}{Jianxin Chang}, \bibinfo{person}{Chen Gao}, \bibinfo{person}{Yu Zheng}, \bibinfo{person}{Yiqun Hui}, \bibinfo{person}{Yanan Niu}, \bibinfo{person}{Yang Song}, \bibinfo{person}{Depeng Jin}, {and} \bibinfo{person}{Yong Li}.} \bibinfo{year}{2021}\natexlab{}.
\newblock \showarticletitle{Sequential recommendation with graph neural networks}. In \bibinfo{booktitle}{\emph{Proceedings of the 44th international ACM SIGIR conference on research and development in information retrieval}}. \bibinfo{pages}{378--387}.
\newblock


\bibitem[\protect\citeauthoryear{Chen, Alhinai, Jiang, Na, and Hao}{Chen et~al\mbox{.}}{2022a}]%
        {chen2022bottleneck}
\bibfield{author}{\bibinfo{person}{Hanqiu Chen}, \bibinfo{person}{Yahya Alhinai}, \bibinfo{person}{Yihan Jiang}, \bibinfo{person}{Eunjee Na}, {and} \bibinfo{person}{Cong Hao}.} \bibinfo{year}{2022}\natexlab{a}.
\newblock \showarticletitle{Bottleneck Analysis of Dynamic Graph Neural Network Inference on CPU and GPU}. In \bibinfo{booktitle}{\emph{2022 IEEE International Symposium on Workload Characterization (IISWC)}}. \bibinfo{pages}{130--145}.
\newblock
\urldef\tempurl%
\url{https://doi.org/10.1109/IISWC55918.2022.00021}
\showDOI{\tempurl}


\bibitem[\protect\citeauthoryear{Chen, Qiu, Li, and Xie}{Chen et~al\mbox{.}}{2022b}]%
        {chen2022graphad}
\bibfield{author}{\bibinfo{person}{Xu Chen}, \bibinfo{person}{Qiu Qiu}, \bibinfo{person}{Changshan Li}, {and} \bibinfo{person}{Kunqing Xie}.} \bibinfo{year}{2022}\natexlab{b}.
\newblock \showarticletitle{GraphAD: a graph neural network for entity-wise multivariate time-series anomaly detection}. In \bibinfo{booktitle}{\emph{Proceedings of the 45th International ACM SIGIR Conference on Research and Development in Information Retrieval}}. \bibinfo{pages}{2297--2302}.
\newblock


\bibitem[\protect\citeauthoryear{Chen, Zhu, Xu, Liu, Xiong, Zhang, and Song}{Chen et~al\mbox{.}}{2021}]%
        {EDGE}
\bibfield{author}{\bibinfo{person}{Xinshi Chen}, \bibinfo{person}{Yan Zhu}, \bibinfo{person}{Haowen Xu}, \bibinfo{person}{Mengyang Liu}, \bibinfo{person}{Liang Xiong}, \bibinfo{person}{Muhan Zhang}, {and} \bibinfo{person}{Le Song}.} \bibinfo{year}{2021}\natexlab{}.
\newblock \showarticletitle{Efficient Dynamic Graph Representation Learning at Scale}.
\newblock \bibinfo{journal}{\emph{arXiv preprint arXiv:2112.07768}} (\bibinfo{year}{2021}).
\newblock


\bibitem[\protect\citeauthoryear{Cong, Kang, Tong, and Mahdavi}{Cong et~al\mbox{.}}{2024}]%
        {cong2024generalization}
\bibfield{author}{\bibinfo{person}{Weilin Cong}, \bibinfo{person}{Jian Kang}, \bibinfo{person}{Hanghang Tong}, {and} \bibinfo{person}{Mehrdad Mahdavi}.} \bibinfo{year}{2024}\natexlab{}.
\newblock \showarticletitle{On the Generalization Capability of Temporal Graph Learning Algorithms: Theoretical Insights and a Simpler Method}.
\newblock \bibinfo{journal}{\emph{arXiv preprint arXiv:2402.16387}} (\bibinfo{year}{2024}).
\newblock


\bibitem[\protect\citeauthoryear{Cong, Zhang, Kang, Yuan, Wu, Zhou, Tong, and Mahdavi}{Cong et~al\mbox{.}}{2023}]%
        {graphmixer}
\bibfield{author}{\bibinfo{person}{Weilin Cong}, \bibinfo{person}{Si Zhang}, \bibinfo{person}{Jian Kang}, \bibinfo{person}{Baichuan Yuan}, \bibinfo{person}{Hao Wu}, \bibinfo{person}{Xin Zhou}, \bibinfo{person}{Hanghang Tong}, {and} \bibinfo{person}{Mehrdad Mahdavi}.} \bibinfo{year}{2023}\natexlab{}.
\newblock \showarticletitle{Do We Really Need Complicated Model Architectures For Temporal Networks?}. In \bibinfo{booktitle}{\emph{{ICLR}}}.
\newblock


\bibitem[\protect\citeauthoryear{Deng, Zhou, Zeng, Xia, Leung, Li, Kannan, and Prasanna}{Deng et~al\mbox{.}}{2024}]%
        {taser}
\bibfield{author}{\bibinfo{person}{Gangda Deng}, \bibinfo{person}{Hongkuan Zhou}, \bibinfo{person}{Hanqing Zeng}, \bibinfo{person}{Yinglong Xia}, \bibinfo{person}{Christopher Leung}, \bibinfo{person}{Jianbo Li}, \bibinfo{person}{Rajgopal Kannan}, {and} \bibinfo{person}{Viktor Prasanna}.} \bibinfo{year}{2024}\natexlab{}.
\newblock \showarticletitle{TASER: Temporal Adaptive Sampling for Fast and Accurate Dynamic Graph Representation Learning}.
\newblock \bibinfo{journal}{\emph{arXiv preprint arXiv:2402.05396}} (\bibinfo{year}{2024}).
\newblock


\bibitem[\protect\citeauthoryear{Gao and Ribeiro}{Gao and Ribeiro}{2022}]%
        {gao2022equivalence}
\bibfield{author}{\bibinfo{person}{Jianfei Gao} {and} \bibinfo{person}{Bruno Ribeiro}.} \bibinfo{year}{2022}\natexlab{}.
\newblock \showarticletitle{On the equivalence between temporal and static equivariant graph representations}. In \bibinfo{booktitle}{\emph{International Conference on Machine Learning}}. PMLR, \bibinfo{pages}{7052--7076}.
\newblock


\bibitem[\protect\citeauthoryear{Gao, Li, Shen, Shao, and Chen}{Gao et~al\mbox{.}}{2024}]%
        {etc}
\bibfield{author}{\bibinfo{person}{Shihong Gao}, \bibinfo{person}{Yiming Li}, \bibinfo{person}{Yanyan Shen}, \bibinfo{person}{Yingxia Shao}, {and} \bibinfo{person}{Lei Chen}.} \bibinfo{year}{2024}\natexlab{}.
\newblock \showarticletitle{{ETC:} Efficient Training of Temporal Graph Neural Networks over Large-scale Dynamic Graphs}.
\newblock \bibinfo{journal}{\emph{Proc. {VLDB} Endow.}} \bibinfo{volume}{17}, \bibinfo{number}{5} (\bibinfo{year}{2024}), \bibinfo{pages}{1060--1072}.
\newblock
\urldef\tempurl%
\url{https://www.vldb.org/pvldb/vol17/p1060-gao.pdf}
\showURL{%
\tempurl}


\bibitem[\protect\citeauthoryear{Gravina and Bacciu}{Gravina and Bacciu}{2024}]%
        {gravina2024deep}
\bibfield{author}{\bibinfo{person}{Alessio Gravina} {and} \bibinfo{person}{Davide Bacciu}.} \bibinfo{year}{2024}\natexlab{}.
\newblock \showarticletitle{Deep learning for dynamic graphs: models and benchmarks}.
\newblock \bibinfo{journal}{\emph{IEEE Transactions on Neural Networks and Learning Systems}} (\bibinfo{year}{2024}).
\newblock


\bibitem[\protect\citeauthoryear{Halfaker, Keyes, Kluver, Thebault-Spieker, Nguyen, Grandprey-Shores, Uduwage, and Warncke-Wang}{Halfaker et~al\mbox{.}}{2015}]%
        {halfaker2015user}
\bibfield{author}{\bibinfo{person}{Aaron Halfaker}, \bibinfo{person}{Os Keyes}, \bibinfo{person}{Daniel Kluver}, \bibinfo{person}{Jacob Thebault-Spieker}, \bibinfo{person}{Tien Nguyen}, \bibinfo{person}{Kate Grandprey-Shores}, \bibinfo{person}{Anuradha Uduwage}, {and} \bibinfo{person}{Morten Warncke-Wang}.} \bibinfo{year}{2015}\natexlab{}.
\newblock \showarticletitle{User session identification based on strong regularities in inter-activity time}. In \bibinfo{booktitle}{\emph{Proceedings of the 24th International Conference on World Wide Web}}. \bibinfo{pages}{410--418}.
\newblock


\bibitem[\protect\citeauthoryear{Hamilton, Ying, and Leskovec}{Hamilton et~al\mbox{.}}{2017}]%
        {sage}
\bibfield{author}{\bibinfo{person}{Will Hamilton}, \bibinfo{person}{Zhitao Ying}, {and} \bibinfo{person}{Jure Leskovec}.} \bibinfo{year}{2017}\natexlab{}.
\newblock \showarticletitle{Inductive representation learning on large graphs}.
\newblock \bibinfo{journal}{\emph{Advances in neural information processing systems}}  \bibinfo{volume}{30} (\bibinfo{year}{2017}).
\newblock


\bibitem[\protect\citeauthoryear{Huang, Jiang, Rao, Zhang, Han, Zhang, Wang, He, Xu, Zhao, et~al\mbox{.}}{Huang et~al\mbox{.}}{2023}]%
        {benchtemp}
\bibfield{author}{\bibinfo{person}{Qiang Huang}, \bibinfo{person}{Jiawei Jiang}, \bibinfo{person}{Xi~Susie Rao}, \bibinfo{person}{Ce Zhang}, \bibinfo{person}{Zhichao Han}, \bibinfo{person}{Zitao Zhang}, \bibinfo{person}{Xin Wang}, \bibinfo{person}{Yongjun He}, \bibinfo{person}{Quanqing Xu}, \bibinfo{person}{Yang Zhao}, {et~al\mbox{.}}} \bibinfo{year}{2023}\natexlab{}.
\newblock \showarticletitle{BenchTemp: A General Benchmark for Evaluating Temporal Graph Neural Networks}.
\newblock \bibinfo{journal}{\emph{arXiv preprint arXiv:2308.16385}} (\bibinfo{year}{2023}).
\newblock


\bibitem[\protect\citeauthoryear{Huang, Poursafaei, Danovitch, Fey, Hu, Rossi, Leskovec, Bronstein, Rabusseau, and Rabbany}{Huang et~al\mbox{.}}{2024}]%
        {tgb}
\bibfield{author}{\bibinfo{person}{Shenyang Huang}, \bibinfo{person}{Farimah Poursafaei}, \bibinfo{person}{Jacob Danovitch}, \bibinfo{person}{Matthias Fey}, \bibinfo{person}{Weihua Hu}, \bibinfo{person}{Emanuele Rossi}, \bibinfo{person}{Jure Leskovec}, \bibinfo{person}{Michael Bronstein}, \bibinfo{person}{Guillaume Rabusseau}, {and} \bibinfo{person}{Reihaneh Rabbany}.} \bibinfo{year}{2024}\natexlab{}.
\newblock \showarticletitle{Temporal graph benchmark for machine learning on temporal graphs}.
\newblock \bibinfo{journal}{\emph{Advances in Neural Information Processing Systems}}  \bibinfo{volume}{36} (\bibinfo{year}{2024}).
\newblock


\bibitem[\protect\citeauthoryear{Jiang, Wang, Cui, Huang, Yang, and Fan}{Jiang et~al\mbox{.}}{2022}]%
        {astgn}
\bibfield{author}{\bibinfo{person}{Wenjian Jiang}, \bibinfo{person}{Xuhong Wang}, \bibinfo{person}{Ping Cui}, \bibinfo{person}{Bin Huang}, \bibinfo{person}{Yupu Yang}, {and} \bibinfo{person}{Qifu Fan}.} \bibinfo{year}{2022}\natexlab{}.
\newblock \showarticletitle{Adaptive Sampling Temporal Graph Network}. In \bibinfo{booktitle}{\emph{2022 4th International Conference on Advances in Computer Technology, Information Science and Communications (CTISC)}}. IEEE, \bibinfo{pages}{1--8}.
\newblock


\bibitem[\protect\citeauthoryear{Jin, Li, and Pan}{Jin et~al\mbox{.}}{2022}]%
        {neurtw}
\bibfield{author}{\bibinfo{person}{Ming Jin}, \bibinfo{person}{Yuan-Fang Li}, {and} \bibinfo{person}{Shirui Pan}.} \bibinfo{year}{2022}\natexlab{}.
\newblock \showarticletitle{Neural temporal walks: Motif-aware representation learning on continuous-time dynamic graphs}.
\newblock \bibinfo{journal}{\emph{Advances in Neural Information Processing Systems}}  \bibinfo{volume}{35} (\bibinfo{year}{2022}), \bibinfo{pages}{19874--19886}.
\newblock


\bibitem[\protect\citeauthoryear{Kazemi, Goel, Eghbali, Ramanan, Sahota, Thakur, Wu, Smyth, Poupart, and Brubaker}{Kazemi et~al\mbox{.}}{2019}]%
        {time2vec}
\bibfield{author}{\bibinfo{person}{Seyed~Mehran Kazemi}, \bibinfo{person}{Rishab Goel}, \bibinfo{person}{Sepehr Eghbali}, \bibinfo{person}{Janahan Ramanan}, \bibinfo{person}{Jaspreet Sahota}, \bibinfo{person}{Sanjay Thakur}, \bibinfo{person}{Stella Wu}, \bibinfo{person}{Cathal Smyth}, \bibinfo{person}{Pascal Poupart}, {and} \bibinfo{person}{Marcus Brubaker}.} \bibinfo{year}{2019}\natexlab{}.
\newblock \showarticletitle{Time2vec: Learning a vector representation of time}.
\newblock \bibinfo{journal}{\emph{arXiv preprint arXiv:1907.05321}} (\bibinfo{year}{2019}).
\newblock


\bibitem[\protect\citeauthoryear{Kazemi, Goel, Jain, Kobyzev, Sethi, Forsyth, and Poupart}{Kazemi et~al\mbox{.}}{2020}]%
        {kazemi2020representation}
\bibfield{author}{\bibinfo{person}{Seyed~Mehran Kazemi}, \bibinfo{person}{Rishab Goel}, \bibinfo{person}{Kshitij Jain}, \bibinfo{person}{Ivan Kobyzev}, \bibinfo{person}{Akshay Sethi}, \bibinfo{person}{Peter Forsyth}, {and} \bibinfo{person}{Pascal Poupart}.} \bibinfo{year}{2020}\natexlab{}.
\newblock \showarticletitle{Representation learning for dynamic graphs: A survey}.
\newblock \bibinfo{journal}{\emph{Journal of Machine Learning Research}} \bibinfo{volume}{21}, \bibinfo{number}{70} (\bibinfo{year}{2020}), \bibinfo{pages}{1--73}.
\newblock


\bibitem[\protect\citeauthoryear{Kumar, Gu, Hoang, Haynes, and Marchetti-Bowick}{Kumar et~al\mbox{.}}{2021}]%
        {9636143}
\bibfield{author}{\bibinfo{person}{Sumit Kumar}, \bibinfo{person}{Yiming Gu}, \bibinfo{person}{Jerrick Hoang}, \bibinfo{person}{Galen~Clark Haynes}, {and} \bibinfo{person}{Micol Marchetti-Bowick}.} \bibinfo{year}{2021}\natexlab{}.
\newblock \showarticletitle{Interaction-Based Trajectory Prediction Over a Hybrid Traffic Graph}. In \bibinfo{booktitle}{\emph{2021 IEEE/RSJ International Conference on Intelligent Robots and Systems (IROS)}}. \bibinfo{pages}{5530--5535}.
\newblock
\urldef\tempurl%
\url{https://doi.org/10.1109/IROS51168.2021.9636143}
\showDOI{\tempurl}


\bibitem[\protect\citeauthoryear{Kumar, Zhang, and Leskovec}{Kumar et~al\mbox{.}}{2019}]%
        {jodie}
\bibfield{author}{\bibinfo{person}{Srijan Kumar}, \bibinfo{person}{Xikun Zhang}, {and} \bibinfo{person}{Jure Leskovec}.} \bibinfo{year}{2019}\natexlab{}.
\newblock \showarticletitle{Predicting dynamic embedding trajectory in temporal interaction networks}. In \bibinfo{booktitle}{\emph{Proceedings of the 25th ACM SIGKDD international conference on knowledge discovery \& data mining}}. \bibinfo{pages}{1269--1278}.
\newblock


\bibitem[\protect\citeauthoryear{Li and Yao}{Li and Yao}{2022}]%
        {li2022corporate}
\bibfield{author}{\bibinfo{person}{Jianing Li} {and} \bibinfo{person}{Xin Yao}.} \bibinfo{year}{2022}\natexlab{}.
\newblock \showarticletitle{Corporate investment prediction using a weighted temporal graph neural network}.
\newblock \bibinfo{journal}{\emph{Wiley Interdisciplinary Reviews: Data Mining and Knowledge Discovery}} \bibinfo{volume}{12}, \bibinfo{number}{6} (\bibinfo{year}{2022}), \bibinfo{pages}{e1472}.
\newblock


\bibitem[\protect\citeauthoryear{Li and Zhu}{Li and Zhu}{2021}]%
        {li2021spatial}
\bibfield{author}{\bibinfo{person}{Mengzhang Li} {and} \bibinfo{person}{Zhanxing Zhu}.} \bibinfo{year}{2021}\natexlab{}.
\newblock \showarticletitle{Spatial-temporal fusion graph neural networks for traffic flow forecasting}. In \bibinfo{booktitle}{\emph{Proceedings of the AAAI conference on artificial intelligence}}, Vol.~\bibinfo{volume}{35}. \bibinfo{pages}{4189--4196}.
\newblock


\bibitem[\protect\citeauthoryear{Li, Shen, Chen, and Yuan}{Li et~al\mbox{.}}{2023a}]%
        {orca}
\bibfield{author}{\bibinfo{person}{Yiming Li}, \bibinfo{person}{Yanyan Shen}, \bibinfo{person}{Lei Chen}, {and} \bibinfo{person}{Mingxuan Yuan}.} \bibinfo{year}{2023}\natexlab{a}.
\newblock \showarticletitle{Orca: Scalable Temporal Graph Neural Network Training with Theoretical Guarantees}.
\newblock \bibinfo{journal}{\emph{Proceedings of the ACM on Management of Data}} \bibinfo{volume}{1}, \bibinfo{number}{1} (\bibinfo{year}{2023}), \bibinfo{pages}{1--27}.
\newblock


\bibitem[\protect\citeauthoryear{Li, Shen, Chen, and Yuan}{Li et~al\mbox{.}}{2023b}]%
        {zebra}
\bibfield{author}{\bibinfo{person}{Yiming Li}, \bibinfo{person}{Yanyan Shen}, \bibinfo{person}{Lei Chen}, {and} \bibinfo{person}{Mingxuan Yuan}.} \bibinfo{year}{2023}\natexlab{b}.
\newblock \showarticletitle{Zebra: When Temporal Graph Neural Networks Meet Temporal Personalized PageRank}.
\newblock \bibinfo{journal}{\emph{Proceedings of the VLDB Endowment}} \bibinfo{volume}{16}, \bibinfo{number}{6} (\bibinfo{year}{2023}), \bibinfo{pages}{1332--1345}.
\newblock


\bibitem[\protect\citeauthoryear{Liu, Liu, Zhao, Tang, and Chen}{Liu et~al\mbox{.}}{2024}]%
        {tpgnn}
\bibfield{author}{\bibinfo{person}{Jie Liu}, \bibinfo{person}{Jiamou Liu}, \bibinfo{person}{Kaiqi Zhao}, \bibinfo{person}{Yanni Tang}, {and} \bibinfo{person}{Wu Chen}.} \bibinfo{year}{2024}\natexlab{}.
\newblock \showarticletitle{TP-GNN: Continuous Dynamic Graph Neural Network for Graph Classification}. In \bibinfo{booktitle}{\emph{2024 IEEE 40th International Conference on Data Engineering (ICDE)}}. IEEE, \bibinfo{pages}{2848--2861}.
\newblock


\bibitem[\protect\citeauthoryear{Luo and Li}{Luo and Li}{2022}]%
        {nat}
\bibfield{author}{\bibinfo{person}{Yuhong Luo} {and} \bibinfo{person}{Pan Li}.} \bibinfo{year}{2022}\natexlab{}.
\newblock \showarticletitle{Neighborhood-aware scalable temporal network representation learning}. In \bibinfo{booktitle}{\emph{Learning on Graphs Conference}}. PMLR, \bibinfo{pages}{1--1}.
\newblock


\bibitem[\protect\citeauthoryear{Marwan, Romano, Thiel, and Kurths}{Marwan et~al\mbox{.}}{2007}]%
        {recur}
\bibfield{author}{\bibinfo{person}{Norbert Marwan}, \bibinfo{person}{M~Carmen Romano}, \bibinfo{person}{Marco Thiel}, {and} \bibinfo{person}{J{\"u}rgen Kurths}.} \bibinfo{year}{2007}\natexlab{}.
\newblock \showarticletitle{Recurrence plots for the analysis of complex systems}.
\newblock \bibinfo{journal}{\emph{Physics reports}} \bibinfo{volume}{438}, \bibinfo{number}{5-6} (\bibinfo{year}{2007}), \bibinfo{pages}{237--329}.
\newblock


\bibitem[\protect\citeauthoryear{Min, Gao, Peng, Wang, Qin, and Fang}{Min et~al\mbox{.}}{2021}]%
        {min2021stgsn}
\bibfield{author}{\bibinfo{person}{Shengjie Min}, \bibinfo{person}{Zhan Gao}, \bibinfo{person}{Jing Peng}, \bibinfo{person}{Liang Wang}, \bibinfo{person}{Ke Qin}, {and} \bibinfo{person}{Bo Fang}.} \bibinfo{year}{2021}\natexlab{}.
\newblock \showarticletitle{Stgsn—a spatial--temporal graph neural network framework for time-evolving social networks}.
\newblock \bibinfo{journal}{\emph{Knowledge-Based Systems}}  \bibinfo{volume}{214} (\bibinfo{year}{2021}), \bibinfo{pages}{106746}.
\newblock


\bibitem[\protect\citeauthoryear{Opsahl and Panzarasa}{Opsahl and Panzarasa}{2009}]%
        {opsahl2009clustering}
\bibfield{author}{\bibinfo{person}{Tore Opsahl} {and} \bibinfo{person}{Pietro Panzarasa}.} \bibinfo{year}{2009}\natexlab{}.
\newblock \showarticletitle{Clustering in weighted networks}.
\newblock \bibinfo{journal}{\emph{Social networks}} \bibinfo{volume}{31}, \bibinfo{number}{2} (\bibinfo{year}{2009}), \bibinfo{pages}{155--163}.
\newblock


\bibitem[\protect\citeauthoryear{Pandey, Li, Hoisie, Li, and Liu}{Pandey et~al\mbox{.}}{2020}]%
        {c-saw}
\bibfield{author}{\bibinfo{person}{Santosh Pandey}, \bibinfo{person}{Lingda Li}, \bibinfo{person}{Adolfy Hoisie}, \bibinfo{person}{Xiaoye~S Li}, {and} \bibinfo{person}{Hang Liu}.} \bibinfo{year}{2020}\natexlab{}.
\newblock \showarticletitle{C-SAW: A framework for graph sampling and random walk on GPUs}. In \bibinfo{booktitle}{\emph{SC20: International Conference for High Performance Computing, Networking, Storage and Analysis}}. IEEE, \bibinfo{pages}{1--15}.
\newblock


\bibitem[\protect\citeauthoryear{Panzarasa, Opsahl, and Carley}{Panzarasa et~al\mbox{.}}{2009}]%
        {panzarasa2009patterns}
\bibfield{author}{\bibinfo{person}{Pietro Panzarasa}, \bibinfo{person}{Tore Opsahl}, {and} \bibinfo{person}{Kathleen~M Carley}.} \bibinfo{year}{2009}\natexlab{}.
\newblock \showarticletitle{Patterns and dynamics of users' behavior and interaction: Network analysis of an online community}.
\newblock \bibinfo{journal}{\emph{Journal of the American Society for Information Science and Technology}} \bibinfo{volume}{60}, \bibinfo{number}{5} (\bibinfo{year}{2009}), \bibinfo{pages}{911--932}.
\newblock


\bibitem[\protect\citeauthoryear{Pareja, Domeniconi, Chen, Ma, Suzumura, Kanezashi, Kaler, Schardl, and Leiserson}{Pareja et~al\mbox{.}}{2020}]%
        {evolvegcn}
\bibfield{author}{\bibinfo{person}{Aldo Pareja}, \bibinfo{person}{Giacomo Domeniconi}, \bibinfo{person}{Jie Chen}, \bibinfo{person}{Tengfei Ma}, \bibinfo{person}{Toyotaro Suzumura}, \bibinfo{person}{Hiroki Kanezashi}, \bibinfo{person}{Tim Kaler}, \bibinfo{person}{Tao Schardl}, {and} \bibinfo{person}{Charles Leiserson}.} \bibinfo{year}{2020}\natexlab{}.
\newblock \showarticletitle{Evolvegcn: Evolving graph convolutional networks for dynamic graphs}. In \bibinfo{booktitle}{\emph{Proceedings of the AAAI conference on artificial intelligence}}, Vol.~\bibinfo{volume}{34}. \bibinfo{pages}{5363--5370}.
\newblock


\bibitem[\protect\citeauthoryear{Poursafaei, Huang, Pelrine, and Rabbany}{Poursafaei et~al\mbox{.}}{2022}]%
        {edgebank}
\bibfield{author}{\bibinfo{person}{Farimah Poursafaei}, \bibinfo{person}{Shenyang Huang}, \bibinfo{person}{Kellin Pelrine}, {and} \bibinfo{person}{Reihaneh Rabbany}.} \bibinfo{year}{2022}\natexlab{}.
\newblock \showarticletitle{Towards better evaluation for dynamic link prediction}.
\newblock \bibinfo{journal}{\emph{Advances in Neural Information Processing Systems}}  \bibinfo{volume}{35} (\bibinfo{year}{2022}), \bibinfo{pages}{32928--32941}.
\newblock


\bibitem[\protect\citeauthoryear{Poursafaei and Rabbany}{Poursafaei and Rabbany}{2023}]%
        {poursafaei2023exhaustive}
\bibfield{author}{\bibinfo{person}{Farimah Poursafaei} {and} \bibinfo{person}{Reihaneh Rabbany}.} \bibinfo{year}{2023}\natexlab{}.
\newblock \showarticletitle{Exhaustive Evaluation of Dynamic Link Prediction}. In \bibinfo{booktitle}{\emph{2023 IEEE International Conference on Data Mining Workshops (ICDMW)}}. IEEE, \bibinfo{pages}{1121--1130}.
\newblock


\bibitem[\protect\citeauthoryear{Rossi, Chamberlain, Frasca, Eynard, Monti, and Bronstein}{Rossi et~al\mbox{.}}{2020}]%
        {tgn}
\bibfield{author}{\bibinfo{person}{Emanuele Rossi}, \bibinfo{person}{Ben Chamberlain}, \bibinfo{person}{Fabrizio Frasca}, \bibinfo{person}{Davide Eynard}, \bibinfo{person}{Federico Monti}, {and} \bibinfo{person}{Michael Bronstein}.} \bibinfo{year}{2020}\natexlab{}.
\newblock \showarticletitle{Temporal Graph Networks for Deep Learning on Dynamic Graphs}. In \bibinfo{booktitle}{\emph{ICML 2020 Workshop on Graph Representation Learning}}.
\newblock


\bibitem[\protect\citeauthoryear{Sankar, Wu, Gou, Zhang, and Yang}{Sankar et~al\mbox{.}}{2020}]%
        {dysat}
\bibfield{author}{\bibinfo{person}{Aravind Sankar}, \bibinfo{person}{Yanhong Wu}, \bibinfo{person}{Liang Gou}, \bibinfo{person}{Wei Zhang}, {and} \bibinfo{person}{Hao Yang}.} \bibinfo{year}{2020}\natexlab{}.
\newblock \showarticletitle{Dysat: Deep neural representation learning on dynamic graphs via self-attention networks}. In \bibinfo{booktitle}{\emph{Proceedings of the 13th international conference on web search and data mining}}. \bibinfo{pages}{519--527}.
\newblock


\bibitem[\protect\citeauthoryear{Tolstikhin, Houlsby, Kolesnikov, Beyer, Zhai, Unterthiner, Yung, Steiner, Keysers, Uszkoreit, et~al\mbox{.}}{Tolstikhin et~al\mbox{.}}{2021}]%
        {mlpmixer}
\bibfield{author}{\bibinfo{person}{Ilya~O Tolstikhin}, \bibinfo{person}{Neil Houlsby}, \bibinfo{person}{Alexander Kolesnikov}, \bibinfo{person}{Lucas Beyer}, \bibinfo{person}{Xiaohua Zhai}, \bibinfo{person}{Thomas Unterthiner}, \bibinfo{person}{Jessica Yung}, \bibinfo{person}{Andreas Steiner}, \bibinfo{person}{Daniel Keysers}, \bibinfo{person}{Jakob Uszkoreit}, {et~al\mbox{.}}} \bibinfo{year}{2021}\natexlab{}.
\newblock \showarticletitle{Mlp-mixer: An all-mlp architecture for vision}.
\newblock \bibinfo{journal}{\emph{Advances in neural information processing systems}} (\bibinfo{year}{2021}), \bibinfo{pages}{24261--24272}.
\newblock


\bibitem[\protect\citeauthoryear{Trivedi, Farajtabar, Biswal, and Zha}{Trivedi et~al\mbox{.}}{2019}]%
        {dyrep}
\bibfield{author}{\bibinfo{person}{Rakshit Trivedi}, \bibinfo{person}{Mehrdad Farajtabar}, \bibinfo{person}{Prasenjeet Biswal}, {and} \bibinfo{person}{Hongyuan Zha}.} \bibinfo{year}{2019}\natexlab{}.
\newblock \showarticletitle{Dyrep: Learning representations over dynamic graphs}. In \bibinfo{booktitle}{\emph{International conference on learning representations}}.
\newblock


\bibitem[\protect\citeauthoryear{Vaswani, Shazeer, Parmar, Uszkoreit, Jones, Gomez, Kaiser, and Polosukhin}{Vaswani et~al\mbox{.}}{2017}]%
        {vaswani2017attention}
\bibfield{author}{\bibinfo{person}{Ashish Vaswani}, \bibinfo{person}{Noam Shazeer}, \bibinfo{person}{Niki Parmar}, \bibinfo{person}{Jakob Uszkoreit}, \bibinfo{person}{Llion Jones}, \bibinfo{person}{Aidan~N Gomez}, \bibinfo{person}{{\L}ukasz Kaiser}, {and} \bibinfo{person}{Illia Polosukhin}.} \bibinfo{year}{2017}\natexlab{}.
\newblock \showarticletitle{Attention is all you need}.
\newblock \bibinfo{journal}{\emph{Advances in neural information processing systems}}  \bibinfo{volume}{30} (\bibinfo{year}{2017}).
\newblock


\bibitem[\protect\citeauthoryear{Wang, Zhang, Zhou, Cui, Fang, Jia, Fang, and Qi}{Wang et~al\mbox{.}}{2021c}]%
        {wang2021temporal}
\bibfield{author}{\bibinfo{person}{Daixin Wang}, \bibinfo{person}{Zhiqiang Zhang}, \bibinfo{person}{Jun Zhou}, \bibinfo{person}{Peng Cui}, \bibinfo{person}{Jingli Fang}, \bibinfo{person}{Quanhui Jia}, \bibinfo{person}{Yanming Fang}, {and} \bibinfo{person}{Yuan Qi}.} \bibinfo{year}{2021}\natexlab{c}.
\newblock \showarticletitle{Temporal-aware graph neural network for credit risk prediction}. In \bibinfo{booktitle}{\emph{Proceedings of the 2021 SIAM International Conference on Data Mining (SDM)}}. SIAM, \bibinfo{pages}{702--710}.
\newblock


\bibitem[\protect\citeauthoryear{Wang, Lyu, Li, Xia, Yang, Wang, Wang, Cui, Yang, Sun, et~al\mbox{.}}{Wang et~al\mbox{.}}{2021b}]%
        {apan}
\bibfield{author}{\bibinfo{person}{Xuhong Wang}, \bibinfo{person}{Ding Lyu}, \bibinfo{person}{Mengjian Li}, \bibinfo{person}{Yang Xia}, \bibinfo{person}{Qi Yang}, \bibinfo{person}{Xinwen Wang}, \bibinfo{person}{Xinguang Wang}, \bibinfo{person}{Ping Cui}, \bibinfo{person}{Yupu Yang}, \bibinfo{person}{Bowen Sun}, {et~al\mbox{.}}} \bibinfo{year}{2021}\natexlab{b}.
\newblock \showarticletitle{Apan: Asynchronous propagation attention network for real-time temporal graph embedding}. In \bibinfo{booktitle}{\emph{Proceedings of the 2021 international conference on management of data}}. \bibinfo{pages}{2628--2638}.
\newblock


\bibitem[\protect\citeauthoryear{Wang, Ma, Wang, Jin, Wang, Tang, Jia, and Yu}{Wang et~al\mbox{.}}{2020}]%
        {wang2020traffic}
\bibfield{author}{\bibinfo{person}{Xiaoyang Wang}, \bibinfo{person}{Yao Ma}, \bibinfo{person}{Yiqi Wang}, \bibinfo{person}{Wei Jin}, \bibinfo{person}{Xin Wang}, \bibinfo{person}{Jiliang Tang}, \bibinfo{person}{Caiyan Jia}, {and} \bibinfo{person}{Jian Yu}.} \bibinfo{year}{2020}\natexlab{}.
\newblock \showarticletitle{Traffic flow prediction via spatial temporal graph neural network}. In \bibinfo{booktitle}{\emph{Proceedings of the web conference 2020}}. \bibinfo{pages}{1082--1092}.
\newblock


\bibitem[\protect\citeauthoryear{Wang, Chang, Liu, Leskovec, and Li}{Wang et~al\mbox{.}}{2021a}]%
        {cawn}
\bibfield{author}{\bibinfo{person}{Yanbang Wang}, \bibinfo{person}{Yen-Yu Chang}, \bibinfo{person}{Yunyu Liu}, \bibinfo{person}{Jure Leskovec}, {and} \bibinfo{person}{Pan Li}.} \bibinfo{year}{2021}\natexlab{a}.
\newblock \showarticletitle{Inductive representation learning in temporal networks via causal anonymous walks}.
\newblock \bibinfo{journal}{\emph{arXiv preprint arXiv:2101.05974}} (\bibinfo{year}{2021}).
\newblock


\bibitem[\protect\citeauthoryear{Wang and Mendis}{Wang and Mendis}{2023}]%
        {tgopt}
\bibfield{author}{\bibinfo{person}{Yufeng Wang} {and} \bibinfo{person}{Charith Mendis}.} \bibinfo{year}{2023}\natexlab{}.
\newblock \showarticletitle{TGOpt: Redundancy-aware optimizations for temporal graph attention networks}. In \bibinfo{booktitle}{\emph{Proceedings of the 28th ACM SIGPLAN Annual Symposium on Principles and Practice of Parallel Programming}}. \bibinfo{pages}{354--368}.
\newblock


\bibitem[\protect\citeauthoryear{Xu, Ruan, K{\"{o}}rpeoglu, Kumar, and Achan}{Xu et~al\mbox{.}}{2020}]%
        {tgat}
\bibfield{author}{\bibinfo{person}{Da Xu}, \bibinfo{person}{Chuanwei Ruan}, \bibinfo{person}{Evren K{\"{o}}rpeoglu}, \bibinfo{person}{Sushant Kumar}, {and} \bibinfo{person}{Kannan Achan}.} \bibinfo{year}{2020}\natexlab{}.
\newblock \showarticletitle{Inductive representation learning on temporal graphs}. In \bibinfo{booktitle}{\emph{{ICLR}}}.
\newblock


\bibitem[\protect\citeauthoryear{Yang, Zhou, Kalander, Huang, and King}{Yang et~al\mbox{.}}{2021}]%
        {yang2021discrete}
\bibfield{author}{\bibinfo{person}{Menglin Yang}, \bibinfo{person}{Min Zhou}, \bibinfo{person}{Marcus Kalander}, \bibinfo{person}{Zengfeng Huang}, {and} \bibinfo{person}{Irwin King}.} \bibinfo{year}{2021}\natexlab{}.
\newblock \showarticletitle{Discrete-time temporal network embedding via implicit hierarchical learning in hyperbolic space}. In \bibinfo{booktitle}{\emph{Proceedings of the 27th ACM SIGKDD Conference on Knowledge Discovery \& Data Mining}}. \bibinfo{pages}{1975--1985}.
\newblock


\bibitem[\protect\citeauthoryear{Yin, Zhang, Wang, Wang, and Li}{Yin et~al\mbox{.}}{2022}]%
        {10.14778/3551793.3551831}
\bibfield{author}{\bibinfo{person}{Haoteng Yin}, \bibinfo{person}{Muhan Zhang}, \bibinfo{person}{Yanbang Wang}, \bibinfo{person}{Jianguo Wang}, {and} \bibinfo{person}{Pan Li}.} \bibinfo{year}{2022}\natexlab{}.
\newblock \showarticletitle{Algorithm and system co-design for efficient subgraph-based graph representation learning}.
\newblock \bibinfo{journal}{\emph{Proc. VLDB Endow.}} \bibinfo{volume}{15}, \bibinfo{number}{11} (\bibinfo{date}{jul} \bibinfo{year}{2022}), \bibinfo{pages}{2788–2796}.
\newblock
\showISSN{2150-8097}
\urldef\tempurl%
\url{https://doi.org/10.14778/3551793.3551831}
\showDOI{\tempurl}


\bibitem[\protect\citeauthoryear{You, Du, and Leskovec}{You et~al\mbox{.}}{2022}]%
        {roland}
\bibfield{author}{\bibinfo{person}{Jiaxuan You}, \bibinfo{person}{Tianyu Du}, {and} \bibinfo{person}{Jure Leskovec}.} \bibinfo{year}{2022}\natexlab{}.
\newblock \showarticletitle{ROLAND: graph learning framework for dynamic graphs}. In \bibinfo{booktitle}{\emph{Proceedings of the 28th ACM SIGKDD conference on knowledge discovery and data mining}}. \bibinfo{pages}{2358--2366}.
\newblock


\bibitem[\protect\citeauthoryear{Yu, Sun, Du, and Lv}{Yu et~al\mbox{.}}{2024}]%
        {dyglib}
\bibfield{author}{\bibinfo{person}{Le Yu}, \bibinfo{person}{Leilei Sun}, \bibinfo{person}{Bowen Du}, {and} \bibinfo{person}{Weifeng Lv}.} \bibinfo{year}{2024}\natexlab{}.
\newblock \showarticletitle{Towards better dynamic graph learning: New architecture and unified library}.
\newblock \bibinfo{journal}{\emph{Advances in Neural Information Processing Systems}}  \bibinfo{volume}{36} (\bibinfo{year}{2024}).
\newblock


\bibitem[\protect\citeauthoryear{Zeng, Jiang, Ding, Li, Hao, and Qiu}{Zeng et~al\mbox{.}}{2021}]%
        {zeng2021hierarchical}
\bibfield{author}{\bibinfo{person}{Xianlin Zeng}, \bibinfo{person}{Yalong Jiang}, \bibinfo{person}{Wenrui Ding}, \bibinfo{person}{Hongguang Li}, \bibinfo{person}{Yafeng Hao}, {and} \bibinfo{person}{Zifeng Qiu}.} \bibinfo{year}{2021}\natexlab{}.
\newblock \showarticletitle{A hierarchical spatio-temporal graph convolutional neural network for anomaly detection in videos}.
\newblock \bibinfo{journal}{\emph{IEEE Transactions on Circuits and Systems for Video Technology}} \bibinfo{volume}{33}, \bibinfo{number}{1} (\bibinfo{year}{2021}), \bibinfo{pages}{200--212}.
\newblock


\bibitem[\protect\citeauthoryear{Zhong, Sheng, Qin, Wang, Gan, and Wu}{Zhong et~al\mbox{.}}{2023}]%
        {gnnflow}
\bibfield{author}{\bibinfo{person}{Yuchen Zhong}, \bibinfo{person}{Guangming Sheng}, \bibinfo{person}{Tianzuo Qin}, \bibinfo{person}{Minjie Wang}, \bibinfo{person}{Quan Gan}, {and} \bibinfo{person}{Chuan Wu}.} \bibinfo{year}{2023}\natexlab{}.
\newblock \showarticletitle{GNNFlow: A Distributed Framework for Continuous Temporal GNN Learning on Dynamic Graphs}.
\newblock \bibinfo{journal}{\emph{arXiv preprint arXiv:2311.17410}} (\bibinfo{year}{2023}).
\newblock


\bibitem[\protect\citeauthoryear{Zhou, Tan, Huang, Zhou, and Wang}{Zhou et~al\mbox{.}}{2021}]%
        {zhou2021temporal}
\bibfield{author}{\bibinfo{person}{Huachi Zhou}, \bibinfo{person}{Qiaoyu Tan}, \bibinfo{person}{Xiao Huang}, \bibinfo{person}{Kaixiong Zhou}, {and} \bibinfo{person}{Xiaoling Wang}.} \bibinfo{year}{2021}\natexlab{}.
\newblock \showarticletitle{Temporal augmented graph neural networks for session-based recommendations}. In \bibinfo{booktitle}{\emph{Proceedings of the 44th International ACM SIGIR conference on research and development in information retrieval}}. \bibinfo{pages}{1798--1802}.
\newblock


\bibitem[\protect\citeauthoryear{Zhou, Zheng, Nisa, Ioannidis, Song, and Karypis}{Zhou et~al\mbox{.}}{2022}]%
        {tgl}
\bibfield{author}{\bibinfo{person}{Hongkuan Zhou}, \bibinfo{person}{Da Zheng}, \bibinfo{person}{Israt Nisa}, \bibinfo{person}{Vasileios Ioannidis}, \bibinfo{person}{Xiang Song}, {and} \bibinfo{person}{George Karypis}.} \bibinfo{year}{2022}\natexlab{}.
\newblock \showarticletitle{TGL: A General Framework for Temporal GNN Training on Billion-Scale Graphs}.
\newblock \bibinfo{journal}{\emph{Proc. VLDB Endow.}} \bibinfo{volume}{15}, \bibinfo{number}{8} (\bibinfo{date}{apr} \bibinfo{year}{2022}), \bibinfo{pages}{1572–1580}.
\newblock
\showISSN{2150-8097}
\urldef\tempurl%
\url{https://doi.org/10.14778/3529337.3529342}
\showDOI{\tempurl}


\bibitem[\protect\citeauthoryear{Zhou, Zheng, Song, Karypis, and Prasanna}{Zhou et~al\mbox{.}}{2023}]%
        {disttgl}
\bibfield{author}{\bibinfo{person}{Hongkuan Zhou}, \bibinfo{person}{Da Zheng}, \bibinfo{person}{Xiang Song}, \bibinfo{person}{George Karypis}, {and} \bibinfo{person}{Viktor Prasanna}.} \bibinfo{year}{2023}\natexlab{}.
\newblock \showarticletitle{DistTGL: Distributed Memory-Based Temporal Graph Neural Network Training}. In \bibinfo{booktitle}{\emph{Proceedings of the International Conference for High Performance Computing, Networking, Storage and Analysis}}. \bibinfo{pages}{1--12}.
\newblock


\end{thebibliography}


\clearpage
\end{document}